\documentclass[sigconf]{acmart}

\AtBeginDocument{%
  }

\copyrightyear{2023}
\acmYear{2023}
\setcopyright{acmlicensed}
\acmConference[MM '23] {Proceedings of the 31st ACM International Conference on Multimedia}{October 29--November 3, 2023}{Ottawa, ON, Canada.}
\acmBooktitle{Proceedings of the 31st ACM International Conference on Multimedia (MM '23), October 29--November 3, 2023, Ottawa, ON, Canada}
\acmPrice{15.00}
\acmISBN{979-8-4007-0108-5/23/10}
\acmDOI{10.1145/3581783.3612255}

\usepackage{booktabs}
\usepackage{amsfonts}
\usepackage{multirow}
\usepackage{balance}

\begin{document}

\title{Taming the Power of Diffusion Models for High-Quality Virtual Try-On with Appearance Flow}

\author{Junhong Gou}
\orcid{0009-0007-6812-9045}
\affiliation{%
  \institution{MoE Key Lab of Artificial Intelligence, Shanghai Jiao Tong University}
  \country{China}
}
\email{goujunhong@sjtu.edu.cn}

\author{Siyu Sun}
\orcid{0009-0003-0861-8788}
\affiliation{%
  \institution{MoE Key Lab of Artificial Intelligence, Shanghai Jiao Tong University}
  \country{China}
}
\email{sunsiyu@sjtu.edu.cn}

\author{Jianfu Zhang}
\orcid{0000-0002-2673-5860}
\authornote{Corresponding authors.}
\affiliation{%
  \institution{Qing Yuan Research Institute, Shanghai Jiao Tong University}
  \country{China}
}
\email{c.sis@sjtu.edu.cn}

\author{Jianlou Si}
\orcid{0000-0002-2029-6588}
\affiliation{%
  \institution{SenseTime Research}
  \country{China}
}
\email{sijianlou@sensetime.com}

\author{Chen Qian}
\orcid{0000-0002-8761-5563}
\affiliation{%
  \institution{SenseTime Research}
  \country{China}
}
\email{qianchen@sensetime.com}

\author{Liqing Zhang}
\orcid{0000-0001-7597-8503}
\authornotemark[1]
\affiliation{%
  \institution{MoE Key Lab of Artificial Intelligence, Shanghai Jiao Tong University}
  \country{China}
}
\email{zhang-lq@cs.sjtu.edu.cn}

\renewcommand{\shortauthors}{Junhong Gou, Siyu Sun, Jianfu Zhang, Jianlou Si, Chen Qian, \& Liqing Zhang}

\begin{abstract}
Virtual try-on is a critical image synthesis task that aims to transfer clothes from one image to another while preserving the details of both humans and clothes. While many existing methods rely on Generative Adversarial Networks (GANs) to achieve this, flaws can still occur, particularly at high resolutions. Recently, the diffusion model has emerged as a promising alternative for generating high-quality images in various applications. However, simply using clothes as a condition for guiding the diffusion model to inpaint is insufficient to maintain the details of the clothes.
To overcome this challenge, we propose an exemplar-based inpainting approach that leverages a warping module to guide the diffusion model's generation effectively. The warping module performs initial processing on the clothes, which helps to preserve the local details of the clothes. We then combine the warped clothes with clothes-agnostic person image and add noise as the input of diffusion model. Additionally, the warped clothes is used as local conditions for each denoising process to ensure that the resulting output retains as much detail as possible. Our approach, namely \textit{\textbf{D}iffusion-based \textbf{C}onditional \textbf{I}npainting for \textbf{V}irtual \textbf{T}ry-\textbf{ON} (DCI-VTON)}, effectively utilizes the power of the diffusion model, and the incorporation of the warping module helps to produce high-quality and realistic virtual try-on results. Experimental results on VITON-HD demonstrate the effectiveness and superiority of our method. Source code and trained models will be publicly released at: https://github.com/bcmi/DCI-VTON-Virtual-Try-On.
\end{abstract}




\begin{teaserfigure}
  \includegraphics[width=\textwidth]{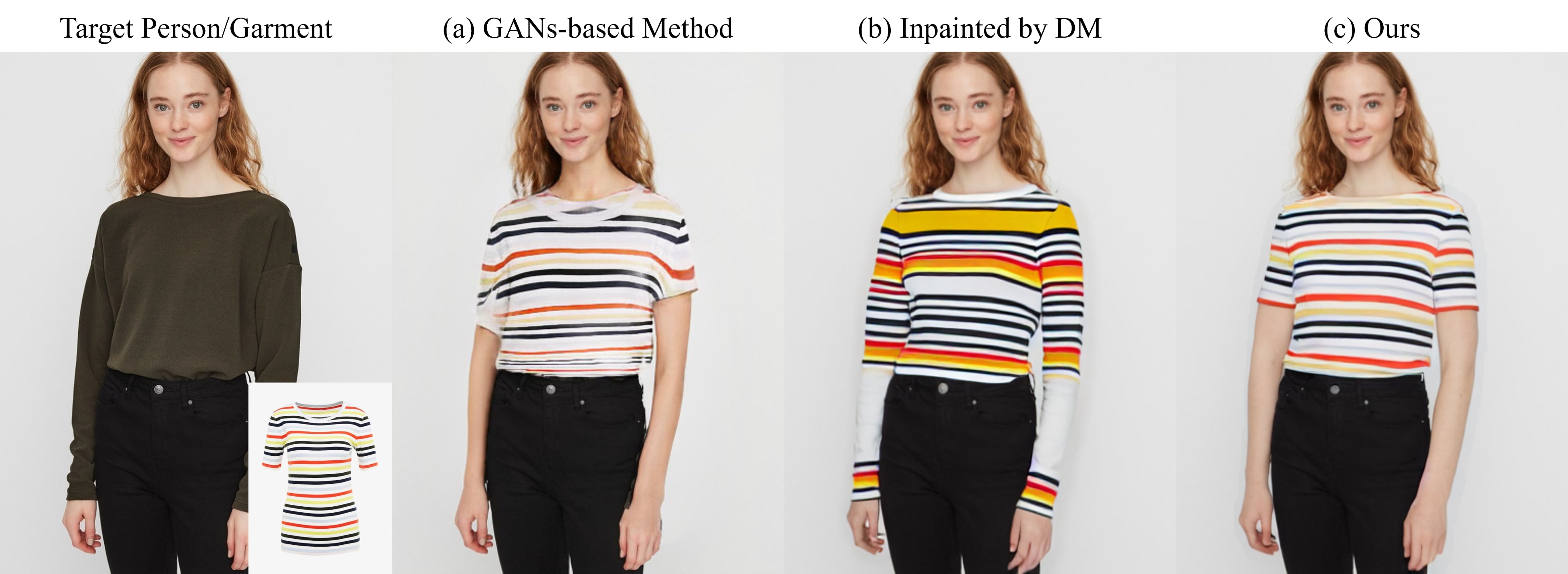}
  \caption{Comparison results of three methods on VITON-HD dataset at 512 $\times$ 384 resolution. It can be seen that our method can generate high-quality results and ensure the restoration of clothes.}
  \label{fig:teaser}
\end{teaserfigure}

\maketitle

\section{Introduction}
Virtual try-on is a prevalently-researched technology that can enhance consumers' shopping experiences. This technique seeks to transfer the clothes in one image to the target person in another image, resulting in a real and plausible composite image. The key point of this task is that, on the presumption that the synthetic results are sufficiently realistic, the textural details of the garment and other character attributes of the target person (\emph{e.g.}, appearance and pose) should be well maintained.

Most of the previous virtual try-on works were based on Generative Adversarial Networks~\cite{goodfellow2014generative} (GANs) in order to generate more realistic pictures. 
To further preserve the details, previous studies~\cite{ge2021parser, han2018viton, minar2020cp, he2022style, yang2020towards, wang2018toward, han2019clothflow} employed an explicit warping module that aligns the target clothes with the human body.
After getting the warped clothes, they fed it into the generator along with the clothes-agnostic image of the person to get the final result.
Based on these, some works~\cite{choi2021viton, lee2022high} additionally expand the task to high-resolution scenarios. 
However, the reliability of such a framework is heavily contingent on the quality of warped garments. Warped garments in low-quality impede faithful generations. 
Furthermore, GANs-based generators inherit the weaknesses of the GAN model, \textit{ i.e.}, convergence heavily depends on the choice of hyperparameters~\cite{gulrajani2017improved, arjovsky2017wasserstein}, and mode drop in the output distribution~\cite{brock2018large, miyato2018spectral}. 
Even though these works have produced some positive outcomes, there are still issues like unrealistic and poor details as shown in Figure \ref{fig:teaser} (a).

More recently, diffusion models~\cite{sohl2015deep, ho2020denoising, song2020denoising, rombach2022high} have gradually emerged and are considered as alternative generative models. 
Compared to GANs, diffusion models can offer desirable qualities, including distribution coverage, a fixed training objective, and scalability~\cite{dhariwal2021diffusion, nichol2021glide}. 
Although the diffusion model has excellent performance in many image generation tasks~\cite{choi2021ilvr, meng2021sdedit, ramesh2022hierarchical, rombach2022high}, virtual try-on remains a very challenging task, for which preserving the detailed features in the reference image (\emph{i.e.}, garment) is critical and essential. 
For our virtual try-on task, a naive method is that we can describe the clothes style through text and then use the mature text-to-image diffusion model framework~\cite{ramesh2022hierarchical, rombach2022high, saharia2022photorealistic} to complete the try-on task. 
However, it is difficult for text to accurately depict some complicated garment texture patterns, resulting in an inability to yield results that are completely consistent with our expectations. 
Recently, \citet{yang2022paint} have proposed a method for exemplar-based image inpainting with diffusion models, which can fill the target region of the source image seamlessly with the objects in the reference image and maintain the overall fidelity and harmonious. 
Similar to this task, we can also regard virtual try-on as an inpainting task.
The primary difference is that the task scene now involves inpainting garments onto humans. 
In this way we can indeed generate high-quality synthetic results, as shown in Figure \ref{fig:teaser} (b). 
However, it is evident that such an approach cannot fully preserve the details of the clothes image, and the clothes style (\emph{e.g.}, color, pattern) is biased. In this example, the color of the clothes and the arrangement of the stripes are completely different from the target clothes.

Motivated by the above points, we propose a virtual try-on framework based on the diffusion model. To fully utilize the diffusion model's powerful generation capabilities while also improving the model's controllability for the try-on task, we divide the entire framework into two major modules, namely the warping module and the refinement module. 
Similar to previous virtual try-on methods~\cite{ge2021parser, lee2022high, he2022style, han2019clothflow}, we predict an appearance flow field in the warping module to fit the clothes to the pose of the target person. 
Then, the warped clothes are directly combined with the image of the person whose torso and arms are masked to get a coarse result. 
This coarse result will be input to our refinement module after adding noise, and an improved result will be obtained after being denoised by the diffusion model. 
A high-quality synthetic result could be produced via such a process, and the powerful generative ability of the diffusion model also ensures that our results will not involve too many artifacts like the previous GANs-based methods. 
After giving an initial guidance of the rough result plus the global conditional guidance of the original clothes image, we also refer to~\cite{yang2022paint} and concatenate the inpaint image and the inpaint mask together as input to control the generation of the diffusion model. Moreover, the warped clothes are combined with inpaint image as local condition to guide each step of the denoising process.
In this way, the issue that the simple inpainting process cannot preserve the details of the clothes has been overcome, as illustrated in Figure \ref{fig:teaser} (c).

To evaluate our proposed method, we conduct extensive experiments on the VITON-HD dataset \cite{choi2021viton} and DressCode dataset~\cite{morelli2022dress}, and compare it with previous works, which proves that our method can achieve excellent performance. Furthermore, we additionally conduct some experiments on virtual try-on task in more complex scenarios on the DeepFashion~\cite{liu2016deepfashion} dataset. 
Specifically, we use another person's clothes as a reference to transfer it to the target person. This task involves the transfer of various human poses, which is more challenging than the scene where template clothes are provided. 

\section{Related Work}
\subsection{Virtual Try-On}
Virtual try-on has always been an appealing research subject since it may significantly enhance the shopping experience of consumers. According to~\cite{feng2022weakly, he2022style}, we can divide the existing virtual try-on technologies into 2D and 3D categories. 3D virtual try-on technology can bring a better user experience, but it relies on 3D parametric human models and unfortunately building a large-scale 3D dataset for training is expensive. Compared with 3D-based methods, image-based virtual try-on, that is, 2D virtual try-on, although not as flexible as 3D (\textit{e.g.}, allowing being viewed with arbitrary views and poses), is more light-weighted and generally more prevalent.

Many previous 2D virtual try-on work~\cite{han2018viton, wang2018toward, yang2020towards, zheng2019virtually, minar2020cp, ge2021disentangled} have used the Thin Plain Spine (TPS) method to flexibly deform clothes to cover the human body. However, TPS can only provide simple deformation processing, which can only roughly migrate the clothes to the target area and cannot handle some larger geometric deformations. In addition, many flow-based methods~\cite{han2019clothflow, ge2021parser, he2022style, bai2022single} have been proposed, they modeled the appearance flow field between clothes and corresponding regions of the human body to better fit the clothes to the person. Most of the previous work was to complete the task of virtual try-on and achieved desirable results under low-resolution conditions. There are also some methods~\cite{choi2021viton, lee2022high} to deal with the virtual try-on task under the high-resolution conditions, which undoubtedly has higher quality requirements in the warping of clothes and the synthesis of images. Most of these works can be divided into two stages. The first stage is the warping stage mentioned earlier, and the second step is the synthesis stage, which is mostly based on GANs. As the resolution increases, it is difficult for these images generated by GANs to retain the characteristics of the clothes, and even the fidelity is significantly decreased with more blurs and artifacts.

The generational capacity of GANs significantly restricts the results of the previous methods. Even if there is a better warping result of clothes, it will still lose a lot of realism when the clothes are combined with the human. It has been proven that the diffusion model is capable of producing high-quality images at high resolutions and has stronger generating capabilities. With the assistance of this innovation, we intend to enhance virtual try-on performance even more.

\subsection{Diffusion Models}
Denoising Diffusion Probabilistic Models (DDPM)~\cite{sohl2015deep, ho2020denoising} has been proposed to generate realistic image from a normal distribution by reversing a gradual noising process. DDPM may generate realistic and diversified images, but its slow sampling speed hinders its broad application. Recently, \citet{song2020denoising} has proposed DDIM to convert the sampling process to a non-Markovian process, enabling faster and deterministic sampling. In order to further reduce the computational complexity and computational resource requirements of the diffusion model, latent diffusion models (LDM)~\cite{rombach2022high} employed a set of frozen encoder-decoder to perform the diffusion and denoising process on the latent space. With the development and maturation of the diffusion model, it has emerged as a formidable competitor to GANs in the field of generation.
 
At the same time, researchers are also exploring how to more effectively control the generation of diffusion models. Text-to-image technology can greatly assist users in their imaginative creations. Many works~\cite{ramesh2022hierarchical, rombach2022high, saharia2022photorealistic} integrate text information as a condition in the denoising process to guide the model to generate images that relate to the text. ILVR~\cite{meng2021sdedit} and SDEdit~\cite{choi2021ilvr} can guide the diffusion model at the spatial level by intervening in the denoising process. More recently, \cite{zhang2023adding, mou2023t2i} have been proposed for easier transfer of diffusion models to different tasks. However, there is still no suitable solution for virtual try-on with diffusion models. In order to depict the various appearances of clothes, it is obviously unrealistic to complete try-on task through the manner of text-to-image. Refer to~\cite{yang2022paint}, we can use the idea of inpainting to complete the try-on task, but this method cannot control the details of inpainting well. To address this issue, we feed the coarse results into the diffusion model for fine-tuning, guiding the generated outcomes effectively. Furthermore, we introduce local conditions in the denoising process, which together with the global conditions to constrain the model generation.

\section{Our Method}
\begin{figure}[t]
    \centering
    \includegraphics[width=1.0\linewidth]{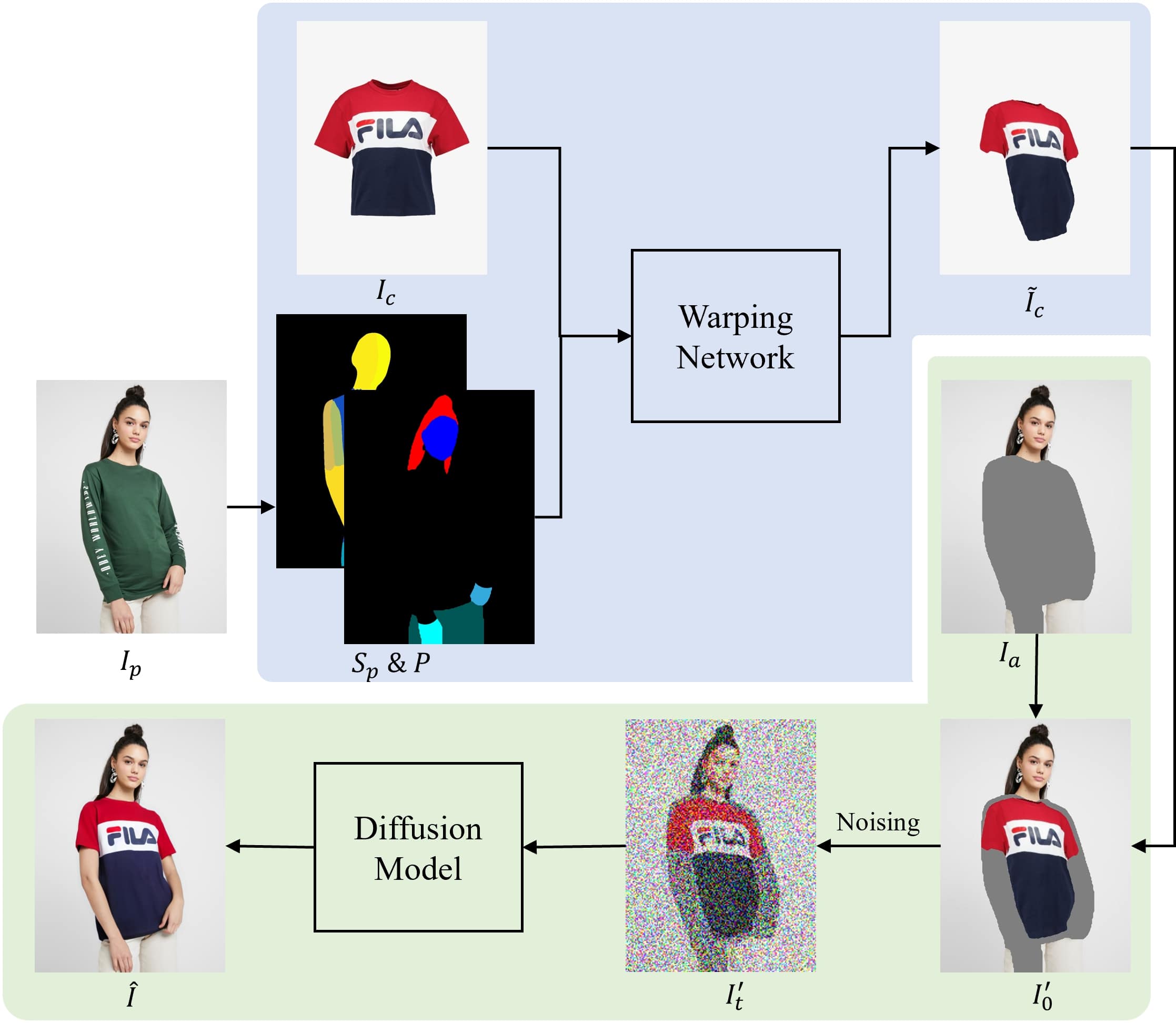}
    \caption{The overview of our method. First, we obtain the segmentation result $S_p$, densepose $P$ and clothes-agnostic $I_a$ of the target person image $I_p$ through preprocessing. The clothes image $I_c$ is roughly aligned to the person by the warping network. Then, We combine $I_a$ and $\tilde{I}_c$ to obtain $I'_0$ and add noise to get $I'_t$ as input to the diffusion model, and the final output $\hat{I}$ produced by denoising $I'_t$. }
    \label{fig:method}
\end{figure}
\begin{figure*}[t]
    \centering
    \includegraphics[width=\linewidth]{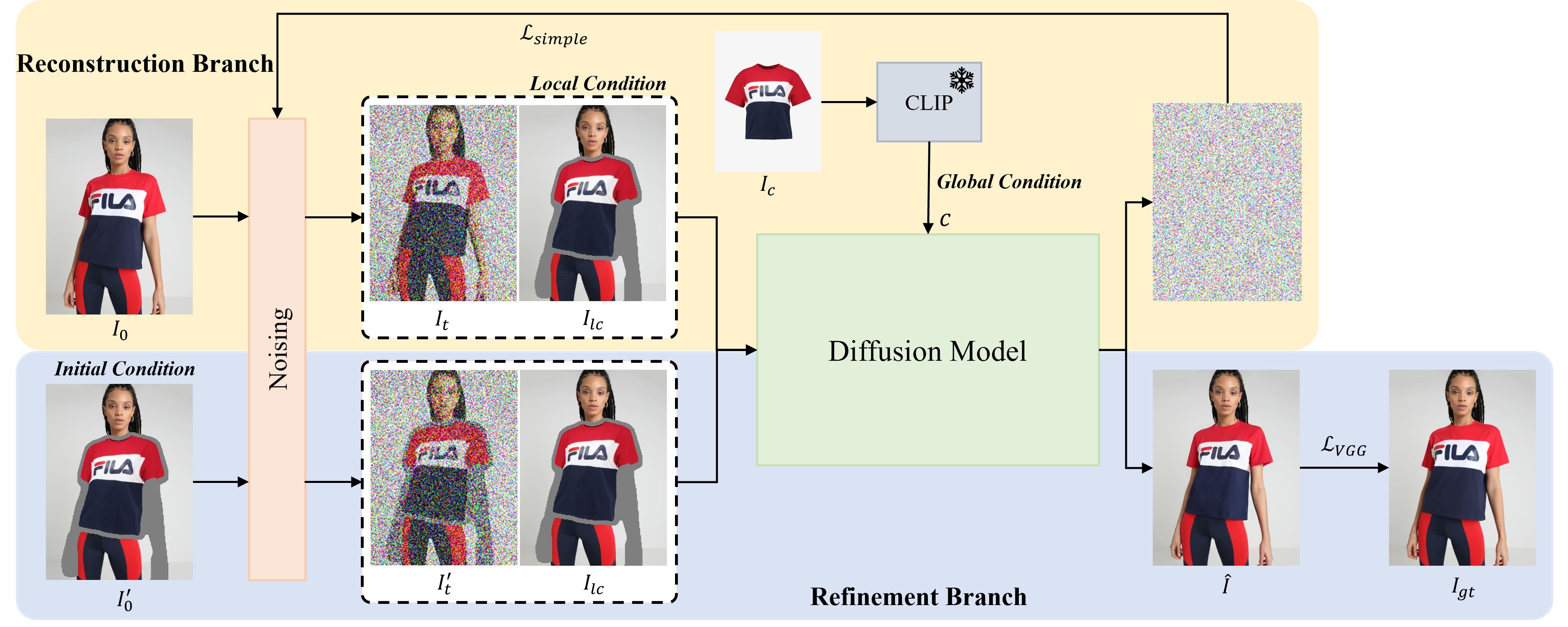}
    \caption{The training pipeline of the diffusion model in our method. There are two branches in our training pipeline: the reconstruction branch above and the refinement branch below. The main difference between them is the inconsistency of the input object and the optimization goal. For better visualization, we show the images corresponding to the variables in the latent space.}
    \label{fig:pipeline}
\end{figure*}
In this work, we seek to employ the diffusion models to accomplish the virtual try-on task in the form of inpainting. Despite the recent remarkable success of text-based image editing, it is still difficult to use mere verbal descriptions to express complex and multiple clothes details. Therefore, it is more practical and feasible to allow users to provide a picture of clothes to achieve a more detailed virtual try-on function.

Formally, given a person image $I_p \in \mathbb{R}^{H\times W\times 3}$ and clothes image $I_c \in \mathbb{R}^{H'\times W'\times 3}$, our goal is to synthesize them into a realistic and plausible image $\hat{I} \in \mathbb{R}^{H\times W\times 3}$, which has the same person attributes as in $I_p$ while retaining the clothes elements from $I_c$. For the mask $m \in \{0,1\}^{H\times W}$ of the area that needs inpainting, in the try-on scene, this area can be fixed as the upper body region of the human body, \emph{i.e.}, the upper part and the arms part. In the synthetic image $\hat{I}$, we hope that the part where $m$ is 0 can only be the same as the $I_p$, and the part where the $m$ is 1 should contain all the elements of $I_c$ and integrate seamlessly with the person. 

To ensure that the clothes in the inpainting region not only maintains most of the original clothes's characteristics but additionally can be ``worn'' by the person in a reasonable manner, we first warp the clothes to align it with the person to create a preliminary composite result, and then refine the inpainting region via the diffusion model. Figure \ref{fig:method} shows the overall process of our method, where the light blue and light green areas represent the processes of warping and refinement respectively. In order to exclude the influence of the clothes worn by the target person of $I_p$ on the succeeding steps, we use person representations extracted from off-the-shelf models~\cite{liang2018look, guler2018densepose} as input. For warping phase, the clothes-agnostic segmentation map $S_p$ is concatenated with densepose $P$, and then, together with the clothes $I_c$, is fed into the warping network to predict an appearance flow field to warp the clothes. The warped clothes $\tilde{I}_c$ and clothes-agnostic person $I_a$ is combined to generate coarse result $I'_0$, which is then noised for subsequent refinement by the diffusion model to get finer results $\hat{I}$.

In the training process, since it is impossible to obtain data pairs of the same person wearing different clothes in the same posture, we use the clothes-agnostic image $I_a$ extracted from $I_p$ and the template image $I_c$ of the clothes on the target person of $I_p$ to reconstruct $I_p$.

\subsection{Warping Network}
There are currently two common methods for warping clothes, namely TPS warping and appearance flow-based warping. The warping method based on the appearance flow has a higher degree of freedom, and correspondingly can adapt to more flexible transformations. The objective of warping network is to predict the dense correspondences between the clothes image and the person image for warping the clothes. Similar to previous works~\cite{han2019clothflow, ge2021parser, lee2022high}, the final flow is obtained by an iterative refinement strategy. This method enables us to capture the long-range correspondence between $I_p$ and $I_c$, allowing us to deal with significant misalignment more effectively.

Specifically, for two kinds of input $I_c$ and $S_p \& P$, we use two symmetrical encoders to extract the feature pyramids $\{E_c\}_{i=1}^N$ and $\{E_p\}_{i=1}^N$. Correspondingly, the flow $F_i$ we predict in each layer will be passed to the next layer for refinement to output $F_{i+1}$ until the final output is obtained. In each layer, the output flow $F_{i-1}$ of the previous layer will first be up-sampled to the same size and warp the corresponding features $E_c^i$, and the result will then correlate with $E_p^i$ to predict the increment of the flow. The final output $F_N \in \mathbb{R}^{H\times W\times 2}$ is a set of
2D coordinate vectors, each of which indicates which pixels in the clothes image $I_c$ should be used to fill the given pixel in the person image $I_p$.

\begin{table*}[t]
  \caption{ Quantitative comparison with baselines. We multiply KID by 100 for better comparison. For User result “a / b”, a is frequency that each method is chosen as the best method for restoring the clothes, and b represents the best generated result.}
  \label{tab:metrics}
  \resizebox{\textwidth}{!} {
  \begin{tabular}{l|cccc|ccccc|cccc}
    \toprule
    \multirow{2}*{Method} & 
         \multicolumn{4}{c}{ 256 $\times$ 192 } & \multicolumn{5}{|c|}{ 512 $\times$ 384 } & \multicolumn{4}{c}{ 1024 $\times$ 768 }  \\
         & LPIPS$\downarrow$ & SSIM$\uparrow$ & FID$\downarrow$ & KID$\downarrow$ & LPIPS$\downarrow$ & SSIM$\uparrow$ & FID$\downarrow$ & KID$\downarrow$ & User$\uparrow$ & LPIPS$\downarrow$ & SSIM$\uparrow$ & FID$\downarrow$ & KID$\downarrow$ \\
    \midrule
    CP-VTON & 0.159 & 0.739 & 30.11 & 2.034 & 0.141 & 0.791 & 30.25 & 4.012 & 0.37\%/0.32\% & 0.158 & 0.786 & 43.28 & 3.762 \\
    VITON-HD & 0.084 & 0.811 & 16.36 & 0.871 & 0.076 & 0.843 & 11.64 & 0.300 & 6.54\%/3.32\% & 0.077 & 0.873 & 11.59 & 0.247 \\
    PF-AFN & 0.089 & 0.863 & 11.49 & 0.319 & 0.082 & 0.858 & 11.30 & 0.283 & 23.78\%/6.93\% & 0.113 & 0.855 & 14.01 & 0.588 \\
    HR-VITON & 0.062 & 0.864 & 9.38 & 0.153 & 0.061 & 0.878 & 9.90 & 0.188 & 27.22\%/7.12\% & 0.065 & \textbf{0.892} & 10.91 & 0.179 \\
    Paint by Example & 0.087 & 0.883 & 9.06 & 0.107 & 0.087 & 0.843 & 10.15 & 0.204 & 0.85\%/15.68\% & 0.157 & 0.821 & 18.12 & 0.782 \\
    Ours & \textbf{0.049} & \textbf{0.906} & \textbf{8.02} & \textbf{0.058} & \textbf{0.043} & \textbf{0.896} & \textbf{8.09} & \textbf{0.028} & \textbf{41.24}\%/\textbf{66.63}\% & \textbf{0.053} & \bf0.892 & \textbf{9.13} & \textbf{0.087} \\
    \bottomrule
  \end{tabular}
  }
\end{table*}

\noindent\textbf{Loss Functions: }
Since the appearance flow is a variable with a high degree of freedom, total-variation (TV) loss can solve this problem well for the smoothness of the final warping result. $\mathcal{L}_{TV}$ can be calculated by the following formula:
\begin{equation}
    \mathcal{L}_{TV} = \sum_{i=1}^N||\nabla F_i ||_1.
\end{equation}
Referring to~\cite{ge2021parser}, we also added a second-order smooth constraint, which is calculated by:
\begin{equation}
    \mathcal{L}_{sec} = \sum_{i=1}^N \sum_t \sum_{\pi \in \mathcal{N}_t} \mathcal{P}(F_i^{t-\pi} + F_i^{t+\pi} - 2F_i^t),
\end{equation}
in which $F_i^t$ indicates the $t$-th point in flow map $F_i$. $\mathcal{N}_t$ indicates the set of horizontal, vertical, and both diagonal neighborhoods around the $t$-th point. $\mathcal{P}$ is generalized charbonnier loss function~\cite{sun2014quantitative}. 
Moreover, for the warped clothes and corresponding warped mask, perceptual loss~\cite{johnson2016perceptual} and L1 loss are used to constrain them to encourage the network to warp the clothes to fit the person’s pose. Formally, $\mathcal{L}_{L1}$ and $\mathcal{L}_{VGG}$ are as follows:
\begin{equation}
    \mathcal{L}_{L1} = \sum_{i=1}^N ||\mathcal{W}(\mathcal{D}_i(M_c), F_i)-\mathcal{D}_i(S_c)||_1,
\end{equation}
\begin{equation}
    \mathcal{L}_{VGG} = \sum_{i=3}^N \sum_{m=1}^5 ||\Phi_m(\mathcal{W}(\mathcal{D}_i(I_c), F_i))-\Phi_m(\mathcal{D}_i(S_c \odot I_p))||_1,
\end{equation}
where $M_c$ and $S_c$ indicate the mask of $I_c$ and clothes mask of $I_p$ respectively. $\mathcal{W}$ represents the warping function, and $\mathcal{D}$ represents the downsampling function. $\Phi_m$ indicates the $m$-th feature map in a VGG-19~\cite{simonyan2014very} network pre-trained on ImageNet~\cite{deng2009imagenet}. 

The total loss function of the entire warping network can be expressed as:
\begin{equation}
    \mathcal{L}_w = \mathcal{L}_{L1} + \lambda_{VGG}\mathcal{L}_{VGG} + \lambda_{TV}\mathcal{L}_{TV} + \lambda_{sec}\mathcal{L}_{sec}.
\end{equation}
where $\lambda_{VGG}, \lambda_{TV}$ and $\lambda_{sec}$ denote the hyper-parameters controlling relative importance between different losses.

\subsection{Diffusion Model}

As indicated in the overview of our strategy in Figure \ref{fig:method}, we intend to apply the diffusion model to refine the coarse synthesis results. To make better use of the initial rough results, we divide the training process into two branches: reconstruction and refinement. Figure \ref{fig:pipeline} depicts our diffusion model training pipeline. During the training process, we will optimize the two branches simultaneously. Intuitively, in the process of optimizing the reconstruction branch, our model can rely on global and local conditions to generate a corresponding real person image. The refinement branch improves the similarity between the prediction results of the model and the rough results by controlling the initial noise. The global condition $c$ indicates the condition extracted by frozen pretrained CLIP~\cite{radford2021learning} image encoder from $I_c$. Due to the cross attention mechanism in LDM~\cite{rombach2022high}, it is easily to use the global attributes of the inpainting object (\emph{e.g.}, shape and pattern category) to guide the generation of the diffusion model, but it is challenging to effectively provide information for some fine-grained attributes (\emph{e.g.}, text, pattern content, and color composition). The lack of details is compensated for by using local conditions. Specifically, we add the warped clothes to the inpainting image $I_a$ as input for each denoising step of the diffusion model. Note that we have not changed the inpainting mask $m$, which means that the clothes in the $I_{lc}$ are only used to provide detailed information, and the final inpainting result will redraw the entire mask area. 
As a result, the clothes in the final composite result might not exactly match its initial warping result. The benefit of this is that it can prevent certain adverse repercussions from poor warping results. Additionally, it can connect the human body part and the clothes part more effectively. In order to make better use of the spatial information contained in the pre-warped clothes and align the final result with the rough result $I'_0$, we also use it as the initial condition, add noise and input it into the diffusion model for refinement.

\noindent\textbf{Reconstruction Branch: }
The reconstruction branch performs similarly to the vanilla diffusion model, which generates realistic images by learning the reverse diffusion process. For the target image $I_0$, we first perform a forward diffusion process, $q(\cdot)$, on it, and gradually add noise to it according to the Markov chain and convert it into a Gaussian distribution. To reduce computational complexity, we employ an latent diffusion model\cite{rombach2022high}, which embeds the images from image space to latent space through a pretrained encoder $\mathcal{E}$ and reconstructs images by a pretrained decoder $\mathcal{D}$. The forward process is performed the latent variable $z_0=\mathcal{E}(I_0)$ at an arbitrary timestamp $t$:
\begin{equation}
\label{eq:noise}
    z_t = \sqrt{\alpha_t} z_0 + \sqrt{1-\alpha_t} \epsilon,
\end{equation}
where $\alpha := \prod_{s=1}^t(1-\beta_s)$ and $\epsilon \sim \mathcal{N}(0, I)$. $\beta$ is a pre-defined variance schedule in $T$ steps. 

Afterwards, we obtain $z_{lc}$ by feeding $I_{lc}$ into the $\mathcal{E}$, and then concatenate them together with the downsampled mask $m$ as the input $\{z_t, z_{lc}, m\}$. During denoising, an enhanced Diffusion UNet~\cite{ronneberger2015u} is used to predict a denoised variant of their input. The global condition $c$ extracted  from $I_c$ is injected into diffusion UNet through cross attention mechanism. So, the objective of this branch is difined as:
\begin{equation}
    \mathcal{L}_{simple}=||\epsilon-\epsilon_\theta(z_t, z_{lc}, m , c, t)||_2.
\end{equation}

\begin{figure*}[t]
    \centering
    \includegraphics[width=\linewidth]{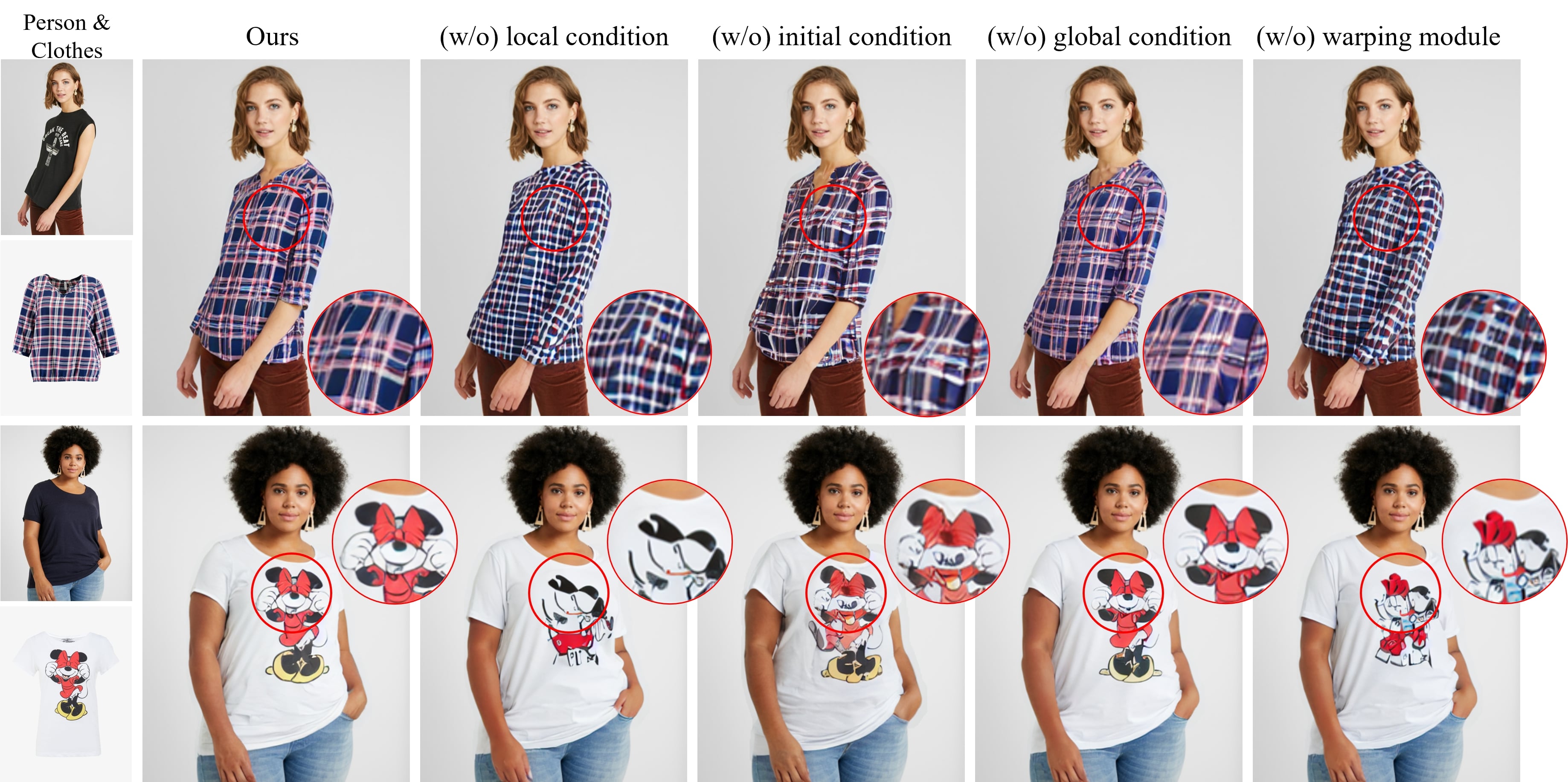}
    \caption{Visual ablation studies of individual components in our approach. }
    \label{fig:ablation}
\end{figure*}

\noindent\textbf{Refinement Branch: }
This branch is based on the rough synthesis result $I'_0$ to inpaint the human body area and deal with the part where the clothes meet the human body, and can also eliminate the negative effects of some inappropriate warping results. Although after the training of the reconstruction branch, the diffusion model can generate a synthetic image that basically restores the characteristics of the clothes under the guidance of local conditions and global conditions, but the lack of spatial guidance makes the generated images unable to fully restore the clothes pattern layout. For example, in the case of a striped clothes, the global condition may prompt the model to build a striped pattern, whereas the local condition adds information such as the thickness and color of the stripe, but these information is insufficient. The initial condition is to further infuse information into the model, such as the arrangement and layout of these stripes. 

Similar to the reconstruct branch, we first employ the encoder $\mathcal{E}$ to extract $z'_0$ from $I'_0$ by $z'_0 = \mathcal{E}(I'_0)$, and then perform forward process on $z'_0$ to get $z'_t$.
Then, $\{z'_t, z_{lc}, m\}$ is fed into the diffusion model for denoising. When the noise $\hat{\epsilon}$ predicted by the model is obtained, according to the Eq.\ref{eq:noise}, we can obtain the refined latent variable $\hat{z}$ after denoising by reverse the equation and the final image result can be recovered such that $\hat{I} = \mathcal{D}(\hat{z})$. After getting $\hat{I}$, we use perceptual loss~\cite{johnson2016perceptual} to optimize it, which can be calculated by:
\begin{equation}
    \mathcal{L}_{VGG} = \sum_{m=1}^5 ||\phi_m(\hat{I}) - \phi_m(I_{gt})||_1.
\end{equation}

Totally, our diffusion model is  trained end-to-end using the following objective function:
\begin{equation}
    \mathcal{L}_d = \mathcal{L}_{simple} + \lambda_{perceptual} \mathcal{L}_{VGG},
\end{equation}
where $\lambda_{perceptual}$ is the hyper-parameter used to balance these two losses.

\section{Experiments}

\subsection{Experiments Setting}

\begin{figure*}
    \centering
    \includegraphics[width=\linewidth]{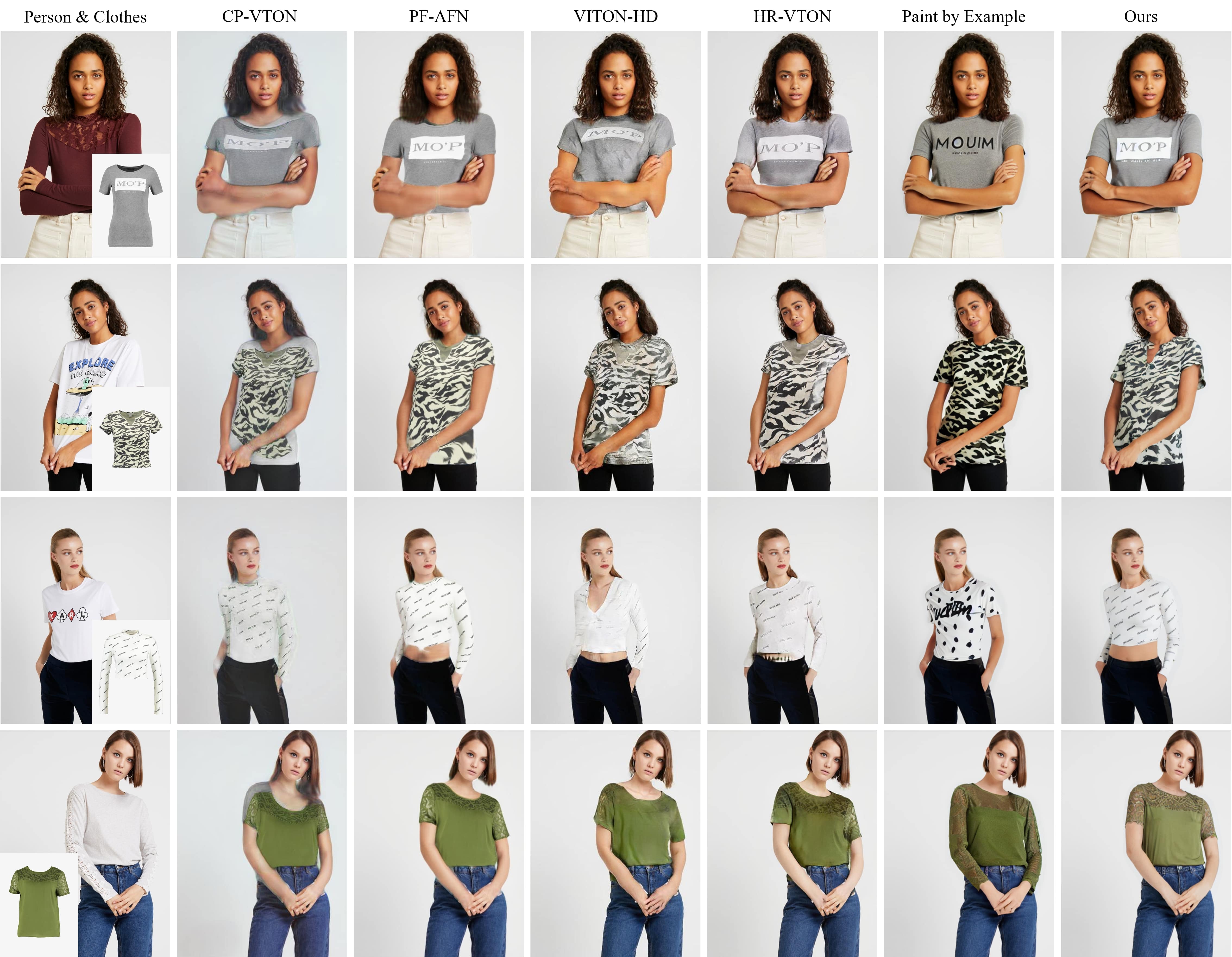}
    \caption{Qualitative comparison with baselines at 512 $\times$ 384 resolution.}
    \label{fig:composite}
\end{figure*}

\noindent\textbf{Datasets: }
Our experiments are mainly carried out on the VITON-HD dataset\cite{choi2021viton}, which contains 13,679 frontal-view woman and top clothes image pairs at the resolution of 1024$\times$768. Following previous work~\cite{choi2021viton, lee2022high}, we split the dataset into a training and a test set with 11,647 and 2,032 pairs respectively, and conduct experiments at there different resolution. Moreover, in order to verify that our method can function in more complicated situations, we also conduct experiments on the DeepFashion dataset~\cite{liu2016deepfashion} and DressCode dataset~\cite{morelli2022dress}, and the experimental results of this part will be provides in the supplementary material.

\noindent\textbf{Evaluation Metrics: }
For the two settings of test, we employ different metrics to evaluate the performance of our method. For the paired setting, which means the clothes image is used to reconstruct person image, we use two widely used metrics: Structural Similarity (SSIM)~\cite{wang2004image} and Learned Perceptual Image Patch Similarity (LPIPS)~\cite{zhang2018unreasonable}. While for the unpaired setting, that is, we need to change the clothes of the person image, we measure Frechet Inception Distance (FID)~\cite{heusel2017gans} and Kernel Inception Distance (KID)~\cite{binkowski2018demystifying}. We consider human perception and include user study for more comprehensive comparison. Specifically, we collect the composite images generated by different methods for 300 pairs randomly selected from the test set at 512 $\times$ 384 resolution. 20 human raters are asked to select the method that restores the most clothes and the method that produces the most realistic results for each test tuple. Then, we report the frequency that each method is selected as the best one in these two aspects.

\noindent\textbf{Implementation Details: }
For the two major modules of our model, the warping module and the refinement module, we train them separately. We train the warping network for 100 epochs with Adam optimizer~\cite{kingma2014adam} for the learning rate of $5\times 10^{-5}$. The hyper-parameters $\lambda_{VGG}, \lambda_{TV}$ and $\lambda_{sec}$ are set as $0.2$, $0.01$ and $6$. Note that, the training of warping module is under the $256\times 192$ resolution. Referring to~\cite{lee2022high}, when inference, we will upsample the predicted appearance flow to the corresponding size.

For the diffusion model, we use KL-regularized autoencoder with latent-space downsampling factor $f=8$. Therefore, the spatial dimension of latent space is $c \times (H/f) \times (W/f)$, where the channel dimension $c$ is $4$. For the denoising UNet, we follow the architecture of~\cite{yang2022paint}. We use AdamW~\cite{loshchilov2017decoupled} optimizer with the learning rate of $1\times 10^{-5}$ and the hyper-parameter $\lambda_{perceptual}$ is set to $1\times 10^{-4}$. We utilize~\cite{yang2022paint} as initialization to provide a strong image prior and basic inpainting ability, and then we train on 2 NVIDIA Tesla A100 GPUs for 40 epochs. During inference, we use PLMS~\cite{liu2022pseudo} sampling method and the number of sampling steps is set to 100.

\subsection{Quantitative Evaluation}
We compare our method with previous virtual try-on methods: CP-VTON~\cite{wang2018toward}, PF-AFN~\cite{ge2021parser}, VITON-HD~\cite{choi2021viton} and HR-VTON~\cite{lee2022high}, and diffusion inpainting method Paint-by-Example~\cite{yang2022paint}. Table \ref{tab:metrics} shows  quantitative comparison with these methods. It can be seen that in the virtual try-on method, HR-VTON achieves state-of-the-art performance at all three resolutions. After being fine-tuned on the VITON-HD dataset, Paint-by-Example also has a very competitive effect. Thanks to the strong image priors embedded in the diffusion model, in the unpaired setting, FID and KID metrics of this method even surpass HR-VTON in some resolution conditions. However, in paired settings, its impact is significantly decreased, owing to the difficulty of preserving most clothes details. In comparison, our method achieves the best results on various metrics and has superior performance in three resolutions. Combining the powerful generation ability of the diffusion model and the strong guidance of our three conditions on the generation process, our model can generate real and natural images while retaining the original clothes to the greatest extent possible.

\subsection{Ablation Study}

\begin{table}[t]
  \caption{Ablation Studies of network components in our model. We multiply KID by 100 for better comparison.}
  \label{tab:ablation}
  \begin{tabular}{lcccc}
    \toprule
    Method & LPIPS$\downarrow$ & SSIM$\uparrow$ & FID$\downarrow$ & KID$\downarrow$ \\
    \midrule
    w/o warping module & 0.054 & 0.891 & 8.13 & 0.034 \\
    w/o global condition & 0.045 & \textbf{0.896} & 8.18 & 0.030 \\
    w/o local condition & 0.065 & 0.888 & 8.14 & 0.032 \\
    w/o initial condition & 0.064 & 0.871 & 10.26 & 0.180 \\
    Ours & \textbf{0.043} & \textbf{0.896} & \textbf{8.09} & \textbf{0.028} \\
    \bottomrule
  \end{tabular}
\end{table}

By taking $512 \times 384$ resolution on VITON-HD dataset as the basic setting, we conduct ablation studies to validate the effectiveness of each component in our network, and the results are shown in Table \ref{tab:ablation}.
First, we explore how much the warping module will affect the subsequent synthesis process (\textbf{w/o warping module}). Referring to~\cite{zhao2021m3d} , we no longer use the warping network to finely warp the clothes, but transform the clothes to a reasonable size and position through the basic affine transformation as the result of the warping and input it into the diffusion model. Specifically, We first center-align the image of the clothes with the inpainting area, and then roughly scale the clothes to fill the inpainting area. This process can be expressed by the following formula:
\begin{equation}
    I_c^{aff} = \left[ \begin{array}{cc}
R & 0 \\
0 & R \\
\end{array} \right] I_c + \left[ \begin{array}{cc}
x^c_{I_a} - x^c_{I_c} \\
y^c_{I_a} - y^c_{I_c} \\
\end{array} \right],
\end{equation}
where $R$ denotes the scale factor computed from the aspect ratio, while $(x^c_{I_a}, y^c_{I_a})$ and $(x^c_{I_c}, y^c_{I_c})$ represent the center of $I_a$ and $I_c$, respectively. It can be shown that the warping module facilitates subsequent synthesis, particularly in complex scenes wherein a person's posture changes significantly and it is difficult to correctly put clothes on the person without pre-warping processing. This also demonstrates that our method is capable of coping with the negative impacts of certain poor warping results.

Afterwards, we explored the influence of the three conditions on the model. First, we remove the global condition (\textbf{w/o global condition}), which means we no longer feed the CLIP features into the network but instead replace them with a learnable variable vector. The global condition among them has the least effect on the model. The primary cause of the limited impact on the results may be that such coarse-grained features are mostly contained by the fine-grained features of other conditions. 
We then try to remove the local condition by using $I_{a}$ instead of $I_{lc}$ in the input of the diffusion model (\textbf{w/o local condition}), only providing guidance outside the inpainting region. It is evident that the lack of local conditions results in some performance reduction.
Following that, we remove the refinement branch, thereby discarding the initial condition (\textbf{w/o initial condition}). Compared with local conditions, the lack of initial conditions has a greater impact on performance, which largely shows that our new refinement branch can make good use of rough results to guide the generated results more accurately. These results demonstrate that the guidance of the three conditions in the process of formation is complementary and indispensable.

In order to more intuitively show the impact of these components on the final result, we visualize them in Figure \ref{fig:ablation}. For such plaid shirts of first row, our full-fledged method can well restore the texture and color on the clothes. In the model that lacks global conditions, in addition to the difference in general color, its results can also restore the characteristics of clothes to a large extent. In the absence of initial conditions, although the stripe arrangement is roughly the same, the distribution and color of each stripe are quite different. In other cases, none of ablative methods can preserve the clothes details well. And for such a meaningful pattern in the second row, only our full-fledged model can preserve it well. In the absence of global conditions, there will still be a certain chromatic aberration. By comparing the results of three and four columns, it can be found that the initial condition is a good complement to the local condition, and it arranges the local conditions spatially. From the results in the last column, it is not difficult to draw the conclusion that pre-warping the clothes can be beneficial in restoring such patterns with practical significance.

\subsection{Qualitative Evaluation}
The composite images produced by various methods on the VITON-HD dataset at 512 $\times$ 384 are exhibited in Figure \ref{fig:composite}. Although some previous virtual try-on methods properly synthesize the human body and clothes, dealing with the interaction between the two is difficult. Paint by Example~\cite{yang2022paint} cannot guarantee that the clothes in the generated results are identical to the given clothes, and there will be texture and pattern differences. It can be seen that our method can generate more realistic and reasonable results than previous methods and can restore the texture characteristics of clothes sufficiently. In the first row, we can see that the previous methods cannot handle the crossed hands of the person well, and our method can cope with such complicated poses well. Similarly, in the second row, the neckline of the clothes and the part where the clothes meet the left hand, our method obtains more realistic results. Moreover, for some transparent materials or hollow styles of clothes, our method can achieve excellent results, as shown in the last row of samples. It is obvious that our method can achieve a more realistic try-on effect for these clothes, such as the mesh style of the clothes in the last row. More examples of composite results and the discussion on limitation of our method are presented in the supplementary materials.

\section{Conclusion}
In this work, we treat the virtual try-on task as an inpainting task and solve it using the diffusion model. In order to allow the diffusion model to better retain the characteristics of the clothes during the inpainting process and improve the authenticity of the generated image, we use a warping network to predict the appearance flow to warp the clothes before inpainting. Under the premise of using the global condition, we add the warped clothes to the input of the diffusion model as the local condition. Meanwhile, a new branch is introduced to assist the model in making better use of the coarse synthesis results obtained in the previous step. The experimental results on the VITON-HD dataset have demonstrated the superiority of our method.

\begin{acks}
 The work was supported by the Shanghai Municipal Science and Technology Major / Key Project, China (Grant No. 20511100300 / 2021SHZDZX0102) and the National Natural Science Foundation of China (Grant No. 62076162).
\end{acks}

\bibliographystyle{ACM-Reference-Format}
\balance
\bibliography{sample-base}


\begin{thebibliography}{49}

%%% the Web address.

\ifx \showCODEN    \undefined \def \showCODEN     #1{\unskip}     \fi
\ifx \showDOI      \undefined \def \showDOI       #1{#1}\fi
\ifx \showISBNx    \undefined \def \showISBNx     #1{\unskip}     \fi
\ifx \showISBNxiii \undefined \def \showISBNxiii  #1{\unskip}     \fi
\ifx \showISSN     \undefined \def \showISSN      #1{\unskip}     \fi
\ifx \showLCCN     \undefined \def \showLCCN      #1{\unskip}     \fi
\ifx \shownote     \undefined \def \shownote      #1{#1}          \fi
\ifx \showarticletitle \undefined \def \showarticletitle #1{#1}   \fi
\ifx \showURL      \undefined \def \showURL       {\relax}        \fi
% invisible to TeX
\providecommand\bibfield[2]{#2}
\providecommand\bibinfo[2]{#2}
\providecommand\natexlab[1]{#1}
\providecommand\showeprint[2][]{arXiv:#2}

\bibitem[Arjovsky et~al\mbox{.}(2017)]%
        {arjovsky2017wasserstein}
\bibfield{author}{\bibinfo{person}{Martin Arjovsky}, \bibinfo{person}{Soumith
  Chintala}, {and} \bibinfo{person}{L{\'e}on Bottou}.}
  \bibinfo{year}{2017}\natexlab{}.
\newblock \showarticletitle{Wasserstein generative adversarial networks}. In
  \bibinfo{booktitle}{\emph{ICML}}.
\newblock


\bibitem[Bai et~al\mbox{.}(2022)]%
        {bai2022single}
\bibfield{author}{\bibinfo{person}{Shuai Bai}, \bibinfo{person}{Huiling Zhou},
  \bibinfo{person}{Zhikang Li}, \bibinfo{person}{Chang Zhou}, {and}
  \bibinfo{person}{Hongxia Yang}.} \bibinfo{year}{2022}\natexlab{}.
\newblock \showarticletitle{Single stage virtual try-on via deformable
  attention flows}. In \bibinfo{booktitle}{\emph{ECCV}}.
\newblock


\bibitem[Bi{\'n}kowski et~al\mbox{.}(2018)]%
        {binkowski2018demystifying}
\bibfield{author}{\bibinfo{person}{Miko{\l}aj Bi{\'n}kowski},
  \bibinfo{person}{Danica~J Sutherland}, \bibinfo{person}{Michael Arbel}, {and}
  \bibinfo{person}{Arthur Gretton}.} \bibinfo{year}{2018}\natexlab{}.
\newblock \showarticletitle{Demystifying mmd gans}.
\newblock \bibinfo{journal}{\emph{arXiv preprint arXiv:1801.01401}}
  (\bibinfo{year}{2018}).
\newblock


\bibitem[Brock et~al\mbox{.}(2018)]%
        {brock2018large}
\bibfield{author}{\bibinfo{person}{Andrew Brock}, \bibinfo{person}{Jeff
  Donahue}, {and} \bibinfo{person}{Karen Simonyan}.}
  \bibinfo{year}{2018}\natexlab{}.
\newblock \showarticletitle{Large scale GAN training for high fidelity natural
  image synthesis}.
\newblock \bibinfo{journal}{\emph{arXiv preprint arXiv:1809.11096}}
  (\bibinfo{year}{2018}).
\newblock


\bibitem[Choi et~al\mbox{.}(2021a)]%
        {choi2021ilvr}
\bibfield{author}{\bibinfo{person}{Jooyoung Choi}, \bibinfo{person}{Sungwon
  Kim}, \bibinfo{person}{Yonghyun Jeong}, \bibinfo{person}{Youngjune Gwon},
  {and} \bibinfo{person}{Sungroh Yoon}.} \bibinfo{year}{2021}\natexlab{a}.
\newblock \showarticletitle{Ilvr: Conditioning method for denoising diffusion
  probabilistic models}.
\newblock \bibinfo{journal}{\emph{arXiv preprint arXiv:2108.02938}}
  (\bibinfo{year}{2021}).
\newblock


\bibitem[Choi et~al\mbox{.}(2021b)]%
        {choi2021viton}
\bibfield{author}{\bibinfo{person}{Seunghwan Choi}, \bibinfo{person}{Sunghyun
  Park}, \bibinfo{person}{Minsoo Lee}, {and} \bibinfo{person}{Jaegul Choo}.}
  \bibinfo{year}{2021}\natexlab{b}.
\newblock \showarticletitle{Viton-hd: High-resolution virtual try-on via
  misalignment-aware normalization}. In \bibinfo{booktitle}{\emph{CVPR}}.
\newblock


\bibitem[Deng et~al\mbox{.}(2009)]%
        {deng2009imagenet}
\bibfield{author}{\bibinfo{person}{Jia Deng}, \bibinfo{person}{Wei Dong},
  \bibinfo{person}{Richard Socher}, \bibinfo{person}{Li-Jia Li},
  \bibinfo{person}{Kai Li}, {and} \bibinfo{person}{Li Fei-Fei}.}
  \bibinfo{year}{2009}\natexlab{}.
\newblock \showarticletitle{Imagenet: A large-scale hierarchical image
  database}. In \bibinfo{booktitle}{\emph{CVPR}}.
\newblock


\bibitem[Dhariwal and Nichol(2021)]%
        {dhariwal2021diffusion}
\bibfield{author}{\bibinfo{person}{Prafulla Dhariwal} {and}
  \bibinfo{person}{Alexander Nichol}.} \bibinfo{year}{2021}\natexlab{}.
\newblock \showarticletitle{Diffusion models beat gans on image synthesis}.
\newblock \bibinfo{journal}{\emph{NeurIPS}} (\bibinfo{year}{2021}).
\newblock


\bibitem[Feng et~al\mbox{.}(2022)]%
        {feng2022weakly}
\bibfield{author}{\bibinfo{person}{Ruili Feng}, \bibinfo{person}{Cheng Ma},
  \bibinfo{person}{Chengji Shen}, \bibinfo{person}{Xin Gao},
  \bibinfo{person}{Zhenjiang Liu}, \bibinfo{person}{Xiaobo Li},
  \bibinfo{person}{Kairi Ou}, \bibinfo{person}{Deli Zhao}, {and}
  \bibinfo{person}{Zheng-Jun Zha}.} \bibinfo{year}{2022}\natexlab{}.
\newblock \showarticletitle{Weakly Supervised High-Fidelity Clothing Model
  Generation}. In \bibinfo{booktitle}{\emph{CVPR}}.
\newblock


\bibitem[Ge et~al\mbox{.}(2021a)]%
        {ge2021disentangled}
\bibfield{author}{\bibinfo{person}{Chongjian Ge}, \bibinfo{person}{Yibing
  Song}, \bibinfo{person}{Yuying Ge}, \bibinfo{person}{Han Yang},
  \bibinfo{person}{Wei Liu}, {and} \bibinfo{person}{Ping Luo}.}
  \bibinfo{year}{2021}\natexlab{a}.
\newblock \showarticletitle{Disentangled cycle consistency for highly-realistic
  virtual try-on}. In \bibinfo{booktitle}{\emph{CVPR}}.
\newblock


\bibitem[Ge et~al\mbox{.}(2021b)]%
        {ge2021parser}
\bibfield{author}{\bibinfo{person}{Yuying Ge}, \bibinfo{person}{Yibing Song},
  \bibinfo{person}{Ruimao Zhang}, \bibinfo{person}{Chongjian Ge},
  \bibinfo{person}{Wei Liu}, {and} \bibinfo{person}{Ping Luo}.}
  \bibinfo{year}{2021}\natexlab{b}.
\newblock \showarticletitle{Parser-free virtual try-on via distilling
  appearance flows}. In \bibinfo{booktitle}{\emph{CVPR}}.
\newblock


\bibitem[Goodfellow et~al\mbox{.}(2014)]%
        {goodfellow2014generative}
\bibfield{author}{\bibinfo{person}{Ian~J. Goodfellow}, \bibinfo{person}{Jean
  Pouget-Abadie}, \bibinfo{person}{Mehdi Mirza}, \bibinfo{person}{Bing Xu},
  \bibinfo{person}{David Warde-Farley}, \bibinfo{person}{Sherjil Ozair},
  \bibinfo{person}{Aaron Courville}, {and} \bibinfo{person}{Yoshua Bengio}.}
  \bibinfo{year}{2014}\natexlab{}.
\newblock \showarticletitle{Generative Adversarial Networks}.
\newblock \bibinfo{journal}{\emph{NeurIPS}} (\bibinfo{year}{2014}).
\newblock


\bibitem[G{\"u}ler et~al\mbox{.}(2018)]%
        {guler2018densepose}
\bibfield{author}{\bibinfo{person}{R{\i}za~Alp G{\"u}ler},
  \bibinfo{person}{Natalia Neverova}, {and} \bibinfo{person}{Iasonas
  Kokkinos}.} \bibinfo{year}{2018}\natexlab{}.
\newblock \showarticletitle{Densepose: Dense human pose estimation in the
  wild}. In \bibinfo{booktitle}{\emph{CVPR}}.
\newblock


\bibitem[Gulrajani et~al\mbox{.}(2017)]%
        {gulrajani2017improved}
\bibfield{author}{\bibinfo{person}{Ishaan Gulrajani}, \bibinfo{person}{Faruk
  Ahmed}, \bibinfo{person}{Martin Arjovsky}, \bibinfo{person}{Vincent
  Dumoulin}, {and} \bibinfo{person}{Aaron~C Courville}.}
  \bibinfo{year}{2017}\natexlab{}.
\newblock \showarticletitle{Improved training of wasserstein gans}.
\newblock \bibinfo{journal}{\emph{NeurIPS}} (\bibinfo{year}{2017}).
\newblock


\bibitem[Han et~al\mbox{.}(2019)]%
        {han2019clothflow}
\bibfield{author}{\bibinfo{person}{Xintong Han}, \bibinfo{person}{Xiaojun Hu},
  \bibinfo{person}{Weilin Huang}, {and} \bibinfo{person}{Matthew~R Scott}.}
  \bibinfo{year}{2019}\natexlab{}.
\newblock \showarticletitle{Clothflow: A flow-based model for clothed person
  generation}. In \bibinfo{booktitle}{\emph{ICCV}}.
\newblock


\bibitem[Han et~al\mbox{.}(2018)]%
        {han2018viton}
\bibfield{author}{\bibinfo{person}{Xintong Han}, \bibinfo{person}{Zuxuan Wu},
  \bibinfo{person}{Zhe Wu}, \bibinfo{person}{Ruichi Yu}, {and}
  \bibinfo{person}{Larry~S Davis}.} \bibinfo{year}{2018}\natexlab{}.
\newblock \showarticletitle{Viton: An image-based virtual try-on network}. In
  \bibinfo{booktitle}{\emph{CVPR}}.
\newblock


\bibitem[He et~al\mbox{.}(2022)]%
        {he2022style}
\bibfield{author}{\bibinfo{person}{Sen He}, \bibinfo{person}{Yi-Zhe Song},
  {and} \bibinfo{person}{Tao Xiang}.} \bibinfo{year}{2022}\natexlab{}.
\newblock \showarticletitle{Style-based global appearance flow for virtual
  try-on}. In \bibinfo{booktitle}{\emph{CVPR}}.
\newblock


\bibitem[Heusel et~al\mbox{.}(2017)]%
        {heusel2017gans}
\bibfield{author}{\bibinfo{person}{Martin Heusel}, \bibinfo{person}{Hubert
  Ramsauer}, \bibinfo{person}{Thomas Unterthiner}, \bibinfo{person}{Bernhard
  Nessler}, {and} \bibinfo{person}{Sepp Hochreiter}.}
  \bibinfo{year}{2017}\natexlab{}.
\newblock \showarticletitle{Gans trained by a two time-scale update rule
  converge to a local nash equilibrium}.
\newblock \bibinfo{journal}{\emph{NeurIPS}} (\bibinfo{year}{2017}).
\newblock


\bibitem[Ho et~al\mbox{.}(2020)]%
        {ho2020denoising}
\bibfield{author}{\bibinfo{person}{Jonathan Ho}, \bibinfo{person}{Ajay Jain},
  {and} \bibinfo{person}{Pieter Abbeel}.} \bibinfo{year}{2020}\natexlab{}.
\newblock \showarticletitle{Denoising diffusion probabilistic models}.
\newblock \bibinfo{journal}{\emph{NeurIPS}} (\bibinfo{year}{2020}).
\newblock


\bibitem[Johnson et~al\mbox{.}(2016)]%
        {johnson2016perceptual}
\bibfield{author}{\bibinfo{person}{Justin Johnson}, \bibinfo{person}{Alexandre
  Alahi}, {and} \bibinfo{person}{Li Fei-Fei}.} \bibinfo{year}{2016}\natexlab{}.
\newblock \showarticletitle{Perceptual losses for real-time style transfer and
  super-resolution}. In \bibinfo{booktitle}{\emph{ECCV}}.
\newblock


\bibitem[Kingma and Ba(2014)]%
        {kingma2014adam}
\bibfield{author}{\bibinfo{person}{Diederik~P Kingma} {and}
  \bibinfo{person}{Jimmy Ba}.} \bibinfo{year}{2014}\natexlab{}.
\newblock \showarticletitle{Adam: A method for stochastic optimization}.
\newblock \bibinfo{journal}{\emph{arXiv preprint arXiv:1412.6980}}
  (\bibinfo{year}{2014}).
\newblock


\bibitem[Lee et~al\mbox{.}(2022)]%
        {lee2022high}
\bibfield{author}{\bibinfo{person}{Sangyun Lee}, \bibinfo{person}{Gyojung Gu},
  \bibinfo{person}{Sunghyun Park}, \bibinfo{person}{Seunghwan Choi}, {and}
  \bibinfo{person}{Jaegul Choo}.} \bibinfo{year}{2022}\natexlab{}.
\newblock \showarticletitle{High-Resolution Virtual Try-On with Misalignment
  and Occlusion-Handled Conditions}. In \bibinfo{booktitle}{\emph{ECCV}}.
\newblock


\bibitem[Liang et~al\mbox{.}(2018)]%
        {liang2018look}
\bibfield{author}{\bibinfo{person}{Xiaodan Liang}, \bibinfo{person}{Ke Gong},
  \bibinfo{person}{Xiaohui Shen}, {and} \bibinfo{person}{Liang Lin}.}
  \bibinfo{year}{2018}\natexlab{}.
\newblock \showarticletitle{Look into person: Joint body parsing \& pose
  estimation network and a new benchmark}.
\newblock \bibinfo{journal}{\emph{TPAMI}} (\bibinfo{year}{2018}).
\newblock


\bibitem[Liu et~al\mbox{.}(2022)]%
        {liu2022pseudo}
\bibfield{author}{\bibinfo{person}{Luping Liu}, \bibinfo{person}{Yi Ren},
  \bibinfo{person}{Zhijie Lin}, {and} \bibinfo{person}{Zhou Zhao}.}
  \bibinfo{year}{2022}\natexlab{}.
\newblock \showarticletitle{Pseudo numerical methods for diffusion models on
  manifolds}.
\newblock \bibinfo{journal}{\emph{arXiv preprint arXiv:2202.09778}}
  (\bibinfo{year}{2022}).
\newblock


\bibitem[Liu et~al\mbox{.}(2016)]%
        {liu2016deepfashion}
\bibfield{author}{\bibinfo{person}{Ziwei Liu}, \bibinfo{person}{Ping Luo},
  \bibinfo{person}{Shi Qiu}, \bibinfo{person}{Xiaogang Wang}, {and}
  \bibinfo{person}{Xiaoou Tang}.} \bibinfo{year}{2016}\natexlab{}.
\newblock \showarticletitle{Deepfashion: Powering robust clothes recognition
  and retrieval with rich annotations}. In \bibinfo{booktitle}{\emph{CVPR}}.
\newblock


\bibitem[Loshchilov and Hutter(2017)]%
        {loshchilov2017decoupled}
\bibfield{author}{\bibinfo{person}{Ilya Loshchilov} {and}
  \bibinfo{person}{Frank Hutter}.} \bibinfo{year}{2017}\natexlab{}.
\newblock \showarticletitle{Decoupled weight decay regularization}.
\newblock \bibinfo{journal}{\emph{arXiv preprint arXiv:1711.05101}}
  (\bibinfo{year}{2017}).
\newblock


\bibitem[Meng et~al\mbox{.}(2021)]%
        {meng2021sdedit}
\bibfield{author}{\bibinfo{person}{Chenlin Meng}, \bibinfo{person}{Yang Song},
  \bibinfo{person}{Jiaming Song}, \bibinfo{person}{Jiajun Wu},
  \bibinfo{person}{Jun-Yan Zhu}, {and} \bibinfo{person}{Stefano Ermon}.}
  \bibinfo{year}{2021}\natexlab{}.
\newblock \showarticletitle{Sdedit: Image synthesis and editing with stochastic
  differential equations}.
\newblock \bibinfo{journal}{\emph{arXiv preprint arXiv:2108.01073}}
  (\bibinfo{year}{2021}).
\newblock


\bibitem[Minar et~al\mbox{.}(2020)]%
        {minar2020cp}
\bibfield{author}{\bibinfo{person}{Matiur~Rahman Minar},
  \bibinfo{person}{Thai~Thanh Tuan}, \bibinfo{person}{Heejune Ahn},
  \bibinfo{person}{Paul Rosin}, {and} \bibinfo{person}{Yu-Kun Lai}.}
  \bibinfo{year}{2020}\natexlab{}.
\newblock \showarticletitle{Cp-vton+: Clothing shape and texture preserving
  image-based virtual try-on}. In \bibinfo{booktitle}{\emph{CVPR Workshops}}.
\newblock


\bibitem[Miyato et~al\mbox{.}(2018)]%
        {miyato2018spectral}
\bibfield{author}{\bibinfo{person}{Takeru Miyato}, \bibinfo{person}{Toshiki
  Kataoka}, \bibinfo{person}{Masanori Koyama}, {and} \bibinfo{person}{Yuichi
  Yoshida}.} \bibinfo{year}{2018}\natexlab{}.
\newblock \showarticletitle{Spectral normalization for generative adversarial
  networks}.
\newblock \bibinfo{journal}{\emph{arXiv preprint arXiv:1802.05957}}
  (\bibinfo{year}{2018}).
\newblock


\bibitem[Morelli et~al\mbox{.}(2022)]%
        {morelli2022dress}
\bibfield{author}{\bibinfo{person}{Davide Morelli}, \bibinfo{person}{Matteo
  Fincato}, \bibinfo{person}{Marcella Cornia}, \bibinfo{person}{Federico
  Landi}, \bibinfo{person}{Fabio Cesari}, {and} \bibinfo{person}{Rita
  Cucchiara}.} \bibinfo{year}{2022}\natexlab{}.
\newblock \showarticletitle{Dress Code: High-Resolution Multi-Category Virtual
  Try-On}. In \bibinfo{booktitle}{\emph{CVPR}}.
\newblock


\bibitem[Mou et~al\mbox{.}(2023)]%
        {mou2023t2i}
\bibfield{author}{\bibinfo{person}{Chong Mou}, \bibinfo{person}{Xintao Wang},
  \bibinfo{person}{Liangbin Xie}, \bibinfo{person}{Jian Zhang},
  \bibinfo{person}{Zhongang Qi}, \bibinfo{person}{Ying Shan}, {and}
  \bibinfo{person}{Xiaohu Qie}.} \bibinfo{year}{2023}\natexlab{}.
\newblock \showarticletitle{T2i-adapter: Learning adapters to dig out more
  controllable ability for text-to-image diffusion models}.
\newblock \bibinfo{journal}{\emph{arXiv preprint arXiv:2302.08453}}
  (\bibinfo{year}{2023}).
\newblock


\bibitem[Nichol et~al\mbox{.}(2021)]%
        {nichol2021glide}
\bibfield{author}{\bibinfo{person}{Alex Nichol}, \bibinfo{person}{Prafulla
  Dhariwal}, \bibinfo{person}{Aditya Ramesh}, \bibinfo{person}{Pranav Shyam},
  \bibinfo{person}{Pamela Mishkin}, \bibinfo{person}{Bob McGrew},
  \bibinfo{person}{Ilya Sutskever}, {and} \bibinfo{person}{Mark Chen}.}
  \bibinfo{year}{2021}\natexlab{}.
\newblock \showarticletitle{Glide: Towards photorealistic image generation and
  editing with text-guided diffusion models}.
\newblock \bibinfo{journal}{\emph{arXiv preprint arXiv:2112.10741}}
  (\bibinfo{year}{2021}).
\newblock


\bibitem[Radford et~al\mbox{.}(2021)]%
        {radford2021learning}
\bibfield{author}{\bibinfo{person}{Alec Radford}, \bibinfo{person}{Jong~Wook
  Kim}, \bibinfo{person}{Chris Hallacy}, \bibinfo{person}{Aditya Ramesh},
  \bibinfo{person}{Gabriel Goh}, \bibinfo{person}{Sandhini Agarwal},
  \bibinfo{person}{Girish Sastry}, \bibinfo{person}{Amanda Askell},
  \bibinfo{person}{Pamela Mishkin}, \bibinfo{person}{Jack Clark},
  {et~al\mbox{.}}} \bibinfo{year}{2021}\natexlab{}.
\newblock \showarticletitle{Learning transferable visual models from natural
  language supervision}. In \bibinfo{booktitle}{\emph{ICML}}.
\newblock


\bibitem[Ramesh et~al\mbox{.}(2022)]%
        {ramesh2022hierarchical}
\bibfield{author}{\bibinfo{person}{Aditya Ramesh}, \bibinfo{person}{Prafulla
  Dhariwal}, \bibinfo{person}{Alex Nichol}, \bibinfo{person}{Casey Chu}, {and}
  \bibinfo{person}{Mark Chen}.} \bibinfo{year}{2022}\natexlab{}.
\newblock \showarticletitle{Hierarchical text-conditional image generation with
  clip latents}.
\newblock \bibinfo{journal}{\emph{arXiv preprint arXiv:2204.06125}}
  (\bibinfo{year}{2022}).
\newblock


\bibitem[Rombach et~al\mbox{.}(2022)]%
        {rombach2022high}
\bibfield{author}{\bibinfo{person}{Robin Rombach}, \bibinfo{person}{Andreas
  Blattmann}, \bibinfo{person}{Dominik Lorenz}, \bibinfo{person}{Patrick
  Esser}, {and} \bibinfo{person}{Bj{\"o}rn Ommer}.}
  \bibinfo{year}{2022}\natexlab{}.
\newblock \showarticletitle{High-resolution image synthesis with latent
  diffusion models}. In \bibinfo{booktitle}{\emph{CVPR}}.
\newblock


\bibitem[Ronneberger et~al\mbox{.}(2015)]%
        {ronneberger2015u}
\bibfield{author}{\bibinfo{person}{Olaf Ronneberger}, \bibinfo{person}{Philipp
  Fischer}, {and} \bibinfo{person}{Thomas Brox}.}
  \bibinfo{year}{2015}\natexlab{}.
\newblock \showarticletitle{U-net: Convolutional networks for biomedical image
  segmentation}. In \bibinfo{booktitle}{\emph{MICCAI}}.
\newblock


\bibitem[Saharia et~al\mbox{.}(2022)]%
        {saharia2022photorealistic}
\bibfield{author}{\bibinfo{person}{Chitwan Saharia}, \bibinfo{person}{William
  Chan}, \bibinfo{person}{Saurabh Saxena}, \bibinfo{person}{Lala Li},
  \bibinfo{person}{Jay Whang}, \bibinfo{person}{Emily~L Denton},
  \bibinfo{person}{Kamyar Ghasemipour}, \bibinfo{person}{Raphael
  Gontijo~Lopes}, \bibinfo{person}{Burcu Karagol~Ayan}, \bibinfo{person}{Tim
  Salimans}, {et~al\mbox{.}}} \bibinfo{year}{2022}\natexlab{}.
\newblock \showarticletitle{Photorealistic text-to-image diffusion models with
  deep language understanding}.
\newblock \bibinfo{journal}{\emph{NeruIPS}} (\bibinfo{year}{2022}).
\newblock


\bibitem[Simonyan and Zisserman(2014)]%
        {simonyan2014very}
\bibfield{author}{\bibinfo{person}{Karen Simonyan} {and}
  \bibinfo{person}{Andrew Zisserman}.} \bibinfo{year}{2014}\natexlab{}.
\newblock \showarticletitle{Very deep convolutional networks for large-scale
  image recognition}.
\newblock \bibinfo{journal}{\emph{arXiv preprint arXiv:1409.1556}}
  (\bibinfo{year}{2014}).
\newblock


\bibitem[Sohl-Dickstein et~al\mbox{.}(2015)]%
        {sohl2015deep}
\bibfield{author}{\bibinfo{person}{Jascha Sohl-Dickstein},
  \bibinfo{person}{Eric Weiss}, \bibinfo{person}{Niru Maheswaranathan}, {and}
  \bibinfo{person}{Surya Ganguli}.} \bibinfo{year}{2015}\natexlab{}.
\newblock \showarticletitle{Deep unsupervised learning using nonequilibrium
  thermodynamics}. In \bibinfo{booktitle}{\emph{ICML}}.
\newblock


\bibitem[Song et~al\mbox{.}(2020)]%
        {song2020denoising}
\bibfield{author}{\bibinfo{person}{Jiaming Song}, \bibinfo{person}{Chenlin
  Meng}, {and} \bibinfo{person}{Stefano Ermon}.}
  \bibinfo{year}{2020}\natexlab{}.
\newblock \showarticletitle{Denoising diffusion implicit models}.
\newblock \bibinfo{journal}{\emph{arXiv preprint arXiv:2010.02502}}
  (\bibinfo{year}{2020}).
\newblock


\bibitem[Sun et~al\mbox{.}(2014)]%
        {sun2014quantitative}
\bibfield{author}{\bibinfo{person}{Deqing Sun}, \bibinfo{person}{Stefan Roth},
  {and} \bibinfo{person}{Michael~J Black}.} \bibinfo{year}{2014}\natexlab{}.
\newblock \showarticletitle{A quantitative analysis of current practices in
  optical flow estimation and the principles behind them}.
\newblock \bibinfo{journal}{\emph{IJCV}} (\bibinfo{year}{2014}).
\newblock


\bibitem[Wang et~al\mbox{.}(2018)]%
        {wang2018toward}
\bibfield{author}{\bibinfo{person}{Bochao Wang}, \bibinfo{person}{Huabin
  Zheng}, \bibinfo{person}{Xiaodan Liang}, \bibinfo{person}{Yimin Chen},
  \bibinfo{person}{Liang Lin}, {and} \bibinfo{person}{Meng Yang}.}
  \bibinfo{year}{2018}\natexlab{}.
\newblock \showarticletitle{Toward characteristic-preserving image-based
  virtual try-on network}. In \bibinfo{booktitle}{\emph{ECCV}}.
\newblock


\bibitem[Wang et~al\mbox{.}(2004)]%
        {wang2004image}
\bibfield{author}{\bibinfo{person}{Zhou Wang}, \bibinfo{person}{Alan~C Bovik},
  \bibinfo{person}{Hamid~R Sheikh}, {and} \bibinfo{person}{Eero~P Simoncelli}.}
  \bibinfo{year}{2004}\natexlab{}.
\newblock \showarticletitle{Image quality assessment: from error visibility to
  structural similarity}.
\newblock \bibinfo{journal}{\emph{TIP}} (\bibinfo{year}{2004}).
\newblock


\bibitem[Yang et~al\mbox{.}(2022)]%
        {yang2022paint}
\bibfield{author}{\bibinfo{person}{Binxin Yang}, \bibinfo{person}{Shuyang Gu},
  \bibinfo{person}{Bo Zhang}, \bibinfo{person}{Ting Zhang},
  \bibinfo{person}{Xuejin Chen}, \bibinfo{person}{Xiaoyan Sun},
  \bibinfo{person}{Dong Chen}, {and} \bibinfo{person}{Fang Wen}.}
  \bibinfo{year}{2022}\natexlab{}.
\newblock \showarticletitle{Paint by Example: Exemplar-based Image Editing with
  Diffusion Models}.
\newblock \bibinfo{journal}{\emph{arXiv preprint arXiv:2211.13227}}
  (\bibinfo{year}{2022}).
\newblock


\bibitem[Yang et~al\mbox{.}(2020)]%
        {yang2020towards}
\bibfield{author}{\bibinfo{person}{Han Yang}, \bibinfo{person}{Ruimao Zhang},
  \bibinfo{person}{Xiaobao Guo}, \bibinfo{person}{Wei Liu},
  \bibinfo{person}{Wangmeng Zuo}, {and} \bibinfo{person}{Ping Luo}.}
  \bibinfo{year}{2020}\natexlab{}.
\newblock \showarticletitle{Towards photo-realistic virtual try-on by
  adaptively generating-preserving image content}. In
  \bibinfo{booktitle}{\emph{CVPR}}.
\newblock


\bibitem[Zhang and Agrawala(2023)]%
        {zhang2023adding}
\bibfield{author}{\bibinfo{person}{Lvmin Zhang} {and} \bibinfo{person}{Maneesh
  Agrawala}.} \bibinfo{year}{2023}\natexlab{}.
\newblock \showarticletitle{Adding conditional control to text-to-image
  diffusion models}.
\newblock \bibinfo{journal}{\emph{arXiv preprint arXiv:2302.05543}}
  (\bibinfo{year}{2023}).
\newblock


\bibitem[Zhang et~al\mbox{.}(2018)]%
        {zhang2018unreasonable}
\bibfield{author}{\bibinfo{person}{Richard Zhang}, \bibinfo{person}{Phillip
  Isola}, \bibinfo{person}{Alexei~A Efros}, \bibinfo{person}{Eli Shechtman},
  {and} \bibinfo{person}{Oliver Wang}.} \bibinfo{year}{2018}\natexlab{}.
\newblock \showarticletitle{The unreasonable effectiveness of deep features as
  a perceptual metric}. In \bibinfo{booktitle}{\emph{CVPR}}.
\newblock


\bibitem[Zhao et~al\mbox{.}(2021)]%
        {zhao2021m3d}
\bibfield{author}{\bibinfo{person}{Fuwei Zhao}, \bibinfo{person}{Zhenyu Xie},
  \bibinfo{person}{Michael Kampffmeyer}, \bibinfo{person}{Haoye Dong},
  \bibinfo{person}{Songfang Han}, \bibinfo{person}{Tianxiang Zheng},
  \bibinfo{person}{Tao Zhang}, {and} \bibinfo{person}{Xiaodan Liang}.}
  \bibinfo{year}{2021}\natexlab{}.
\newblock \showarticletitle{M3d-vton: A monocular-to-3d virtual try-on
  network}. In \bibinfo{booktitle}{\emph{ICCV}}.
\newblock


\bibitem[Zheng et~al\mbox{.}(2019)]%
        {zheng2019virtually}
\bibfield{author}{\bibinfo{person}{Na Zheng}, \bibinfo{person}{Xuemeng Song},
  \bibinfo{person}{Zhaozheng Chen}, \bibinfo{person}{Linmei Hu},
  \bibinfo{person}{Da Cao}, {and} \bibinfo{person}{Liqiang Nie}.}
  \bibinfo{year}{2019}\natexlab{}.
\newblock \showarticletitle{Virtually trying on new clothing with arbitrary
  poses}. In \bibinfo{booktitle}{\emph{ACM MM}}.
\newblock


\end{thebibliography}



\begin{thebibliography}{9}

%%% the Web address.

\ifx \showCODEN    \undefined \def \showCODEN     #1{\unskip}     \fi
\ifx \showDOI      \undefined \def \showDOI       #1{#1}\fi
\ifx \showISBNx    \undefined \def \showISBNx     #1{\unskip}     \fi
\ifx \showISBNxiii \undefined \def \showISBNxiii  #1{\unskip}     \fi
\ifx \showISSN     \undefined \def \showISSN      #1{\unskip}     \fi
\ifx \showLCCN     \undefined \def \showLCCN      #1{\unskip}     \fi
\ifx \shownote     \undefined \def \shownote      #1{#1}          \fi
\ifx \showarticletitle \undefined \def \showarticletitle #1{#1}   \fi
\ifx \showURL      \undefined \def \showURL       {\relax}        \fi
% invisible to TeX
\providecommand\bibfield[2]{#2}
\providecommand\bibinfo[2]{#2}
\providecommand\natexlab[1]{#1}
\providecommand\showeprint[2][]{arXiv:#2}

\bibitem[Choi et~al\mbox{.}(2021)]%
        {choi2021viton}
\bibfield{author}{\bibinfo{person}{Seunghwan Choi}, \bibinfo{person}{Sunghyun
  Park}, \bibinfo{person}{Minsoo Lee}, {and} \bibinfo{person}{Jaegul Choo}.}
  \bibinfo{year}{2021}\natexlab{}.
\newblock \showarticletitle{Viton-hd: High-resolution virtual try-on via
  misalignment-aware normalization}. In \bibinfo{booktitle}{\emph{CVPR}}.
\newblock


\bibitem[Ge et~al\mbox{.}(2021)]%
        {ge2021parser}
\bibfield{author}{\bibinfo{person}{Yuying Ge}, \bibinfo{person}{Yibing Song},
  \bibinfo{person}{Ruimao Zhang}, \bibinfo{person}{Chongjian Ge},
  \bibinfo{person}{Wei Liu}, {and} \bibinfo{person}{Ping Luo}.}
  \bibinfo{year}{2021}\natexlab{}.
\newblock \showarticletitle{Parser-free virtual try-on via distilling
  appearance flows}. In \bibinfo{booktitle}{\emph{CVPR}}.
\newblock


\bibitem[Lee et~al\mbox{.}(2022)]%
        {lee2022high}
\bibfield{author}{\bibinfo{person}{Sangyun Lee}, \bibinfo{person}{Gyojung Gu},
  \bibinfo{person}{Sunghyun Park}, \bibinfo{person}{Seunghwan Choi}, {and}
  \bibinfo{person}{Jaegul Choo}.} \bibinfo{year}{2022}\natexlab{}.
\newblock \showarticletitle{High-Resolution Virtual Try-On with Misalignment
  and Occlusion-Handled Conditions}. In \bibinfo{booktitle}{\emph{ECCV}}.
\newblock


\bibitem[Liu et~al\mbox{.}(2016)]%
        {liu2016deepfashion}
\bibfield{author}{\bibinfo{person}{Ziwei Liu}, \bibinfo{person}{Ping Luo},
  \bibinfo{person}{Shi Qiu}, \bibinfo{person}{Xiaogang Wang}, {and}
  \bibinfo{person}{Xiaoou Tang}.} \bibinfo{year}{2016}\natexlab{}.
\newblock \showarticletitle{Deepfashion: Powering robust clothes recognition
  and retrieval with rich annotations}. In \bibinfo{booktitle}{\emph{CVPR}}.
\newblock


\bibitem[Morelli et~al\mbox{.}(2022)]%
        {morelli2022dress}
\bibfield{author}{\bibinfo{person}{Davide Morelli}, \bibinfo{person}{Matteo
  Fincato}, \bibinfo{person}{Marcella Cornia}, \bibinfo{person}{Federico
  Landi}, \bibinfo{person}{Fabio Cesari}, {and} \bibinfo{person}{Rita
  Cucchiara}.} \bibinfo{year}{2022}\natexlab{}.
\newblock \showarticletitle{Dress Code: High-Resolution Multi-Category Virtual
  Try-On}. In \bibinfo{booktitle}{\emph{CVPR}}.
\newblock


\bibitem[Rombach et~al\mbox{.}(2022)]%
        {rombach2022high}
\bibfield{author}{\bibinfo{person}{Robin Rombach}, \bibinfo{person}{Andreas
  Blattmann}, \bibinfo{person}{Dominik Lorenz}, \bibinfo{person}{Patrick
  Esser}, {and} \bibinfo{person}{Bj{\"o}rn Ommer}.}
  \bibinfo{year}{2022}\natexlab{}.
\newblock \showarticletitle{High-resolution image synthesis with latent
  diffusion models}. In \bibinfo{booktitle}{\emph{CVPR}}.
\newblock


\bibitem[Wang et~al\mbox{.}(2018)]%
        {wang2018toward}
\bibfield{author}{\bibinfo{person}{Bochao Wang}, \bibinfo{person}{Huabin
  Zheng}, \bibinfo{person}{Xiaodan Liang}, \bibinfo{person}{Yimin Chen},
  \bibinfo{person}{Liang Lin}, {and} \bibinfo{person}{Meng Yang}.}
  \bibinfo{year}{2018}\natexlab{}.
\newblock \showarticletitle{Toward characteristic-preserving image-based
  virtual try-on network}. In \bibinfo{booktitle}{\emph{ECCV}}.
\newblock


\bibitem[Yang et~al\mbox{.}(2022)]%
        {yang2022paint}
\bibfield{author}{\bibinfo{person}{Binxin Yang}, \bibinfo{person}{Shuyang Gu},
  \bibinfo{person}{Bo Zhang}, \bibinfo{person}{Ting Zhang},
  \bibinfo{person}{Xuejin Chen}, \bibinfo{person}{Xiaoyan Sun},
  \bibinfo{person}{Dong Chen}, {and} \bibinfo{person}{Fang Wen}.}
  \bibinfo{year}{2022}\natexlab{}.
\newblock \showarticletitle{Paint by Example: Exemplar-based Image Editing with
  Diffusion Models}.
\newblock \bibinfo{journal}{\emph{arXiv preprint arXiv:2211.13227}}
  (\bibinfo{year}{2022}).
\newblock


\bibitem[Zhu et~al\mbox{.}(2019)]%
        {zhu2019progressive}
\bibfield{author}{\bibinfo{person}{Zhen Zhu}, \bibinfo{person}{Tengteng Huang},
  \bibinfo{person}{Baoguang Shi}, \bibinfo{person}{Miao Yu},
  \bibinfo{person}{Bofei Wang}, {and} \bibinfo{person}{Xiang Bai}.}
  \bibinfo{year}{2019}\natexlab{}.
\newblock \showarticletitle{Progressive pose attention transfer for person
  image generation}. In \bibinfo{booktitle}{\emph{Proceedings of the IEEE/CVF
  Conference on Computer Vision and Pattern Recognition}}.
\newblock


\end{thebibliography}

\end{document}

\title{Supplementary for Taming the Power of Diffusion Models for High-Quality Virtual Try-On with Appearance Flow}

\author{Junhong Gou}
\orcid{0009-0007-6812-9045}
\affiliation{%
  \institution{MoE Key Lab of Artificial Intelligence, Shanghai Jiao Tong University}
  \country{China}
}
\email{goujunhong@sjtu.edu.cn}

\author{Siyu Sun}
\orcid{0009-0003-0861-8788}
\affiliation{%
  \institution{MoE Key Lab of Artificial Intelligence, Shanghai Jiao Tong University}
  \country{China}
}
\email{sunsiyu@sjtu.edu.cn}

\author{Jianfu Zhang}
\orcid{0000-0002-2673-5860}
\authornote{Corresponding authors.}
\affiliation{%
  \institution{Qing Yuan Research Institute, Shanghai Jiao Tong University}
  \country{China}
}
\email{c.sis@sjtu.edu.cn}

\author{Jianlou Si}
\orcid{0000-0002-2029-6588}
\affiliation{%
  \institution{SenseTime Research}
  \country{China}
}
\email{sijianlou@sensetime.com}

\author{Chen Qian}
\orcid{0000-0002-8761-5563}
\affiliation{%
  \institution{SenseTime Research}
  \country{China}
}
\email{qianchen@sensetime.com}

\author{Liqing Zhang}
\orcid{0000-0001-7597-8503}
\authornotemark[1]
\affiliation{%
  \institution{MoE Key Lab of Artificial Intelligence, Shanghai Jiao Tong University}
  \country{China}
}
\email{zhang-lq@cs.sjtu.edu.cn}

\renewcommand{\shortauthors}{Junhong Gou, Siyu Sun, Jianfu Zhang, Jianlou Si, Chen Qian, \& Liqing Zhang}

\maketitle
\appendix

In this document, we provide additional materials to supplement our main text. In Appendix \ref{sec:result}, we show more qualitative comparison results on the VITON-HD~\cite{choi2021viton} dataset. Additionally, we perform experiments on DressCode~\cite{morelli2022dress} and DeepFashion~\cite{liu2016deepfashion} datasets, and more qualitative results would be shown. Then, we compare our proposed approach to text-to-image based inpainting approach in Appendix \ref{sec:text}. Finally, we show failure cases generated by our method and discuss the limitations of our method in Appendix \ref{sec:limit}.

\section{More Qualitative Results}
\label{sec:result}
\subsection{Results on VITON-HD}

\begin{figure*}
    \centering
    \includegraphics[width=\linewidth]{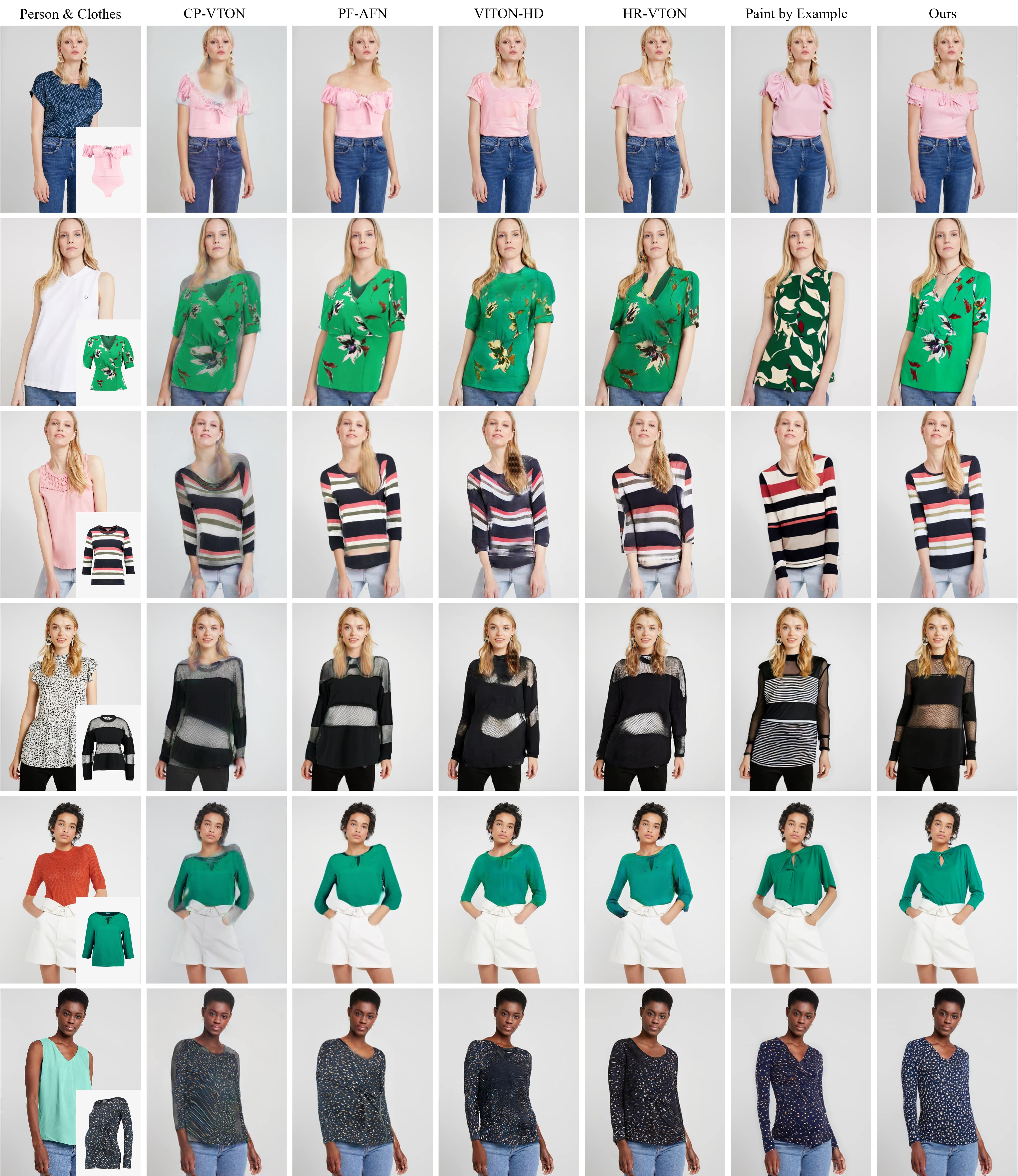}
    \caption{Qualitative comparison of different methods on VITON-HD dataset.}
    \label{fig:visual1}
\end{figure*}

\begin{figure*}
    \centering
    \includegraphics[width=\linewidth]{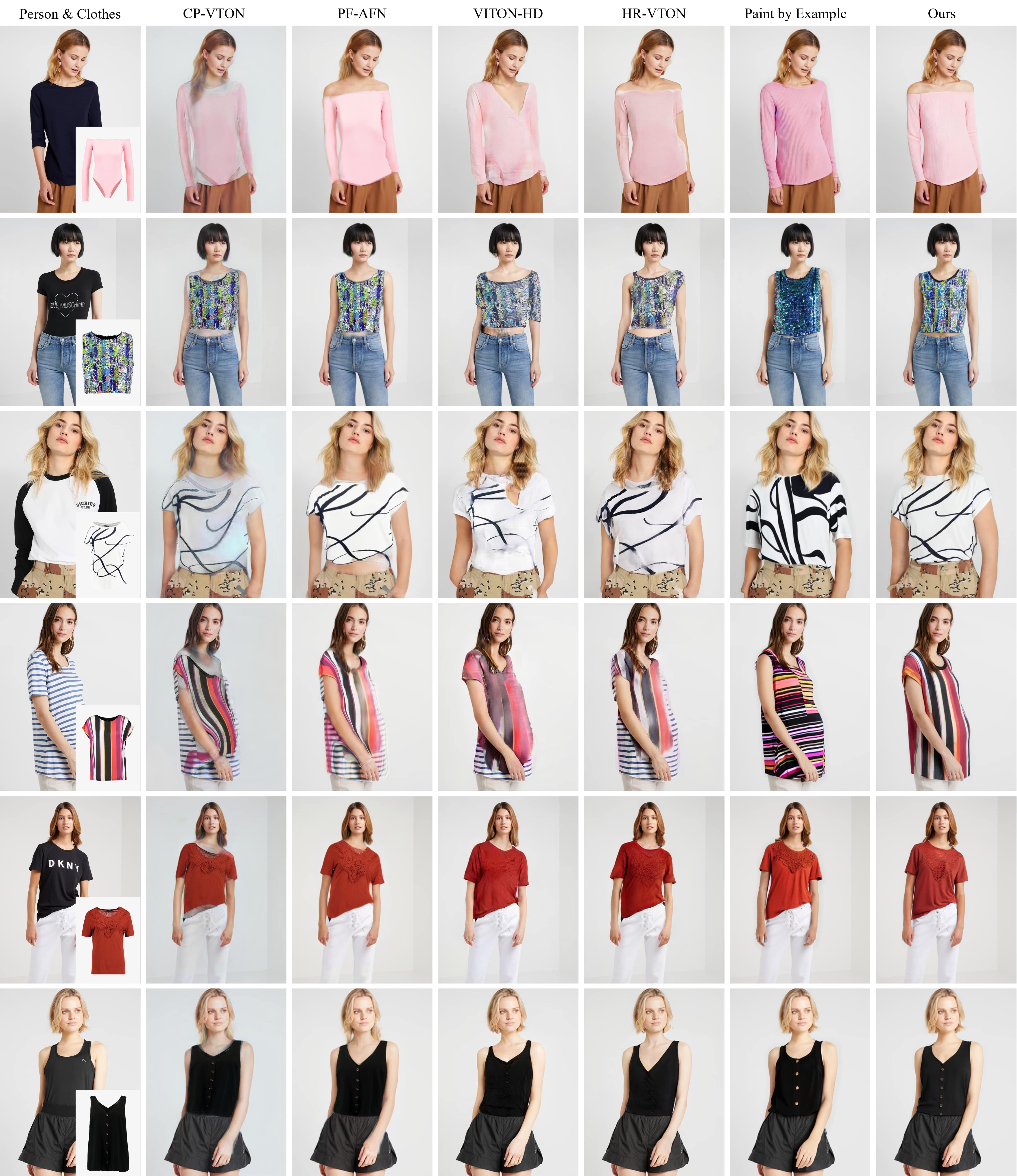}
    \caption{Qualitative comparison of different methods on VITON-HD dataset.}
    \label{fig:visual2}
\end{figure*}

\begin{figure*}
    \centering
    \includegraphics[width=\linewidth]{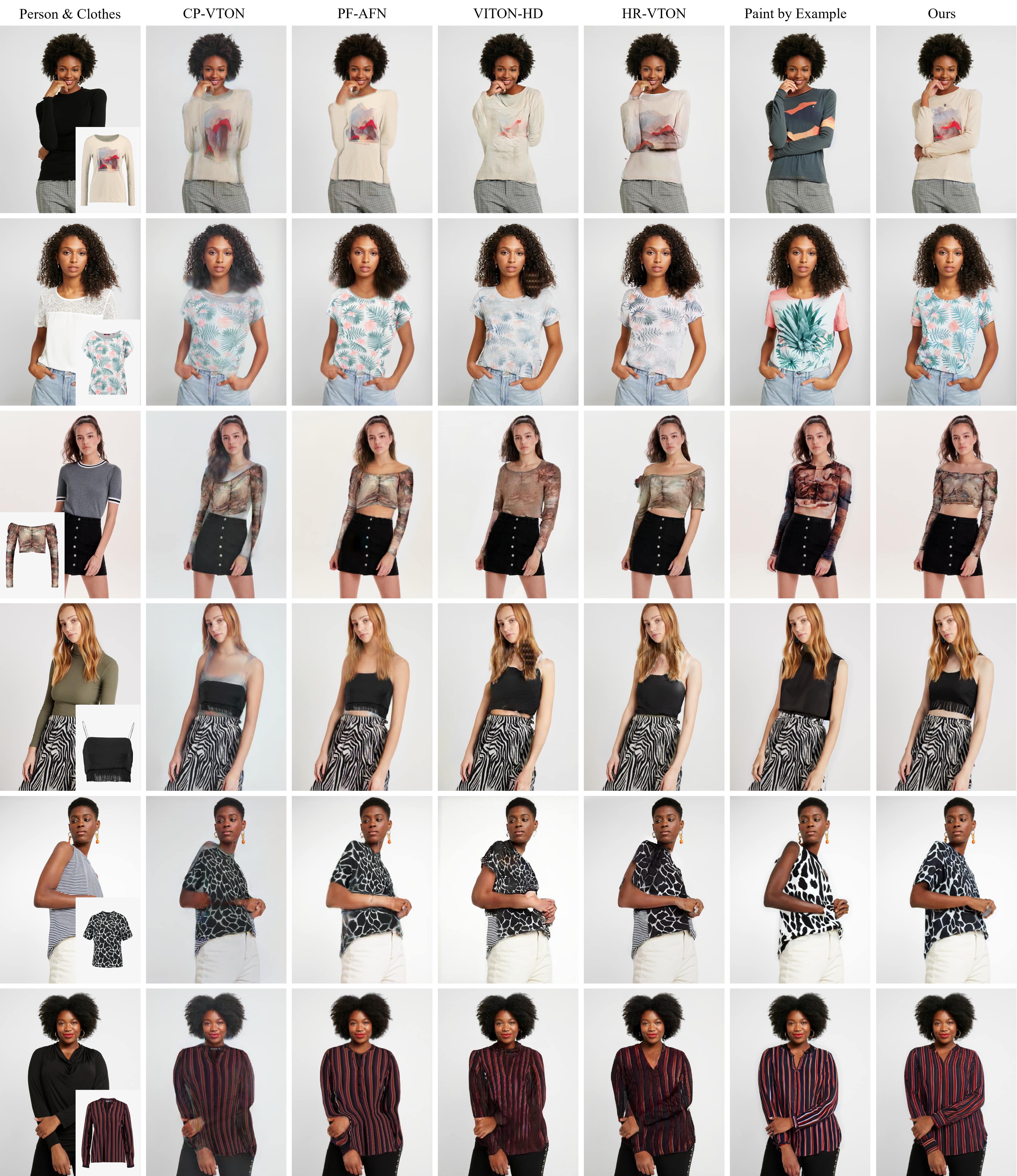}
    \caption{Qualitative comparison of different methods on VITON-HD dataset.}
    \label{fig:visual3}
\end{figure*}

In this section, we show more composite images produced by various methods on VITON-HD dataset in Figure \ref{fig:visual1}, \ref{fig:visual2} and \ref{fig:visual3}. The previous methods include CP-VTON~\cite{wang2018toward}, PF-AFN~\cite{ge2021parser}, VITON-HD~\cite{choi2021viton}, HR-VTON~\cite{lee2022high}, and diffusion inpainting method Paint-by-Example~\cite{yang2022paint}. It is evident that our method outperforms the previous methods in terms of the characteristic restoration of clothes and the authenticity of synthesized pictures. Other methods generally suffer from issues with insufficient restoration of clothing and excessively blurry, unrealistic results.

In the third row of Figure \ref{fig:visual1}, it is difficult for many previous methods to maintain the fluidity of the stripes on the clothes, especially in the part that is in contact with the hair. Our approach effectively addresses these issues while preserving the original stripe layout. As for the clothes made of tulle in the fourth row, due to the image prior contained in the diffusion model, we can better restore the characteristics of such materials. In addition, when facing the densely spotted clothing texture in the last row, we can see that most of these spots in the results generated by the previous method are blurred or disappear. In Figure \ref{fig:visual2}, for people standing sideways like in the fourth row, the previous method cannot handle such a situation well. In contrast, our method can get a more reasonable result. Additionally, in the fifth row our method can more effectively produce stacking wrinkles on the clothing to improve realism. 
Moreover, in the 1st, 5th and 6th rows of Figure \ref{fig:visual3}, the texture of the clothing is blocked by the arms of the person. Other methods cannot reasonably guarantee the pattern layout of the clothing, but our method can handle such occlusion more effectively.

\subsection{Results on DressCode}

\begin{table*}[t]
  \caption{ Quantitative comparison with baselines on DressCode dataset.}
  \label{tab:dresscode}
  \begin{tabular}{l|cc|cc|cc}
    \toprule
    \multirow{2}*{Method} & 
         \multicolumn{2}{c}{ DressCode-Upper } & \multicolumn{2}{|c|}{ DressCode-Lower } & \multicolumn{2}{c}{ DressCode-Dresses }  \\
         & LPIPS$\downarrow$ & FID$\downarrow$ &  LPIPS$\downarrow$ & FID$\downarrow$ & LPIPS$\downarrow$ & FID$\downarrow$ \\
    \midrule
    PF-AFN & 0.0380 & 14.32 & 0.0445 & 18.32 & 0.0758 & 13.59 \\
    HR-VTON & 0.0635 & 16.86 & 0.0811 & 22.81 & 0.1132 & 16.12 \\
    Ours & \textbf{0.0301} & \textbf{10.82} & \textbf{0.0348} & \textbf{12.34} & \textbf{0.0681} & \textbf{12.25} \\
    \bottomrule
  \end{tabular}
\end{table*}

\begin{figure*}
    \centering
    \includegraphics[width=\linewidth]{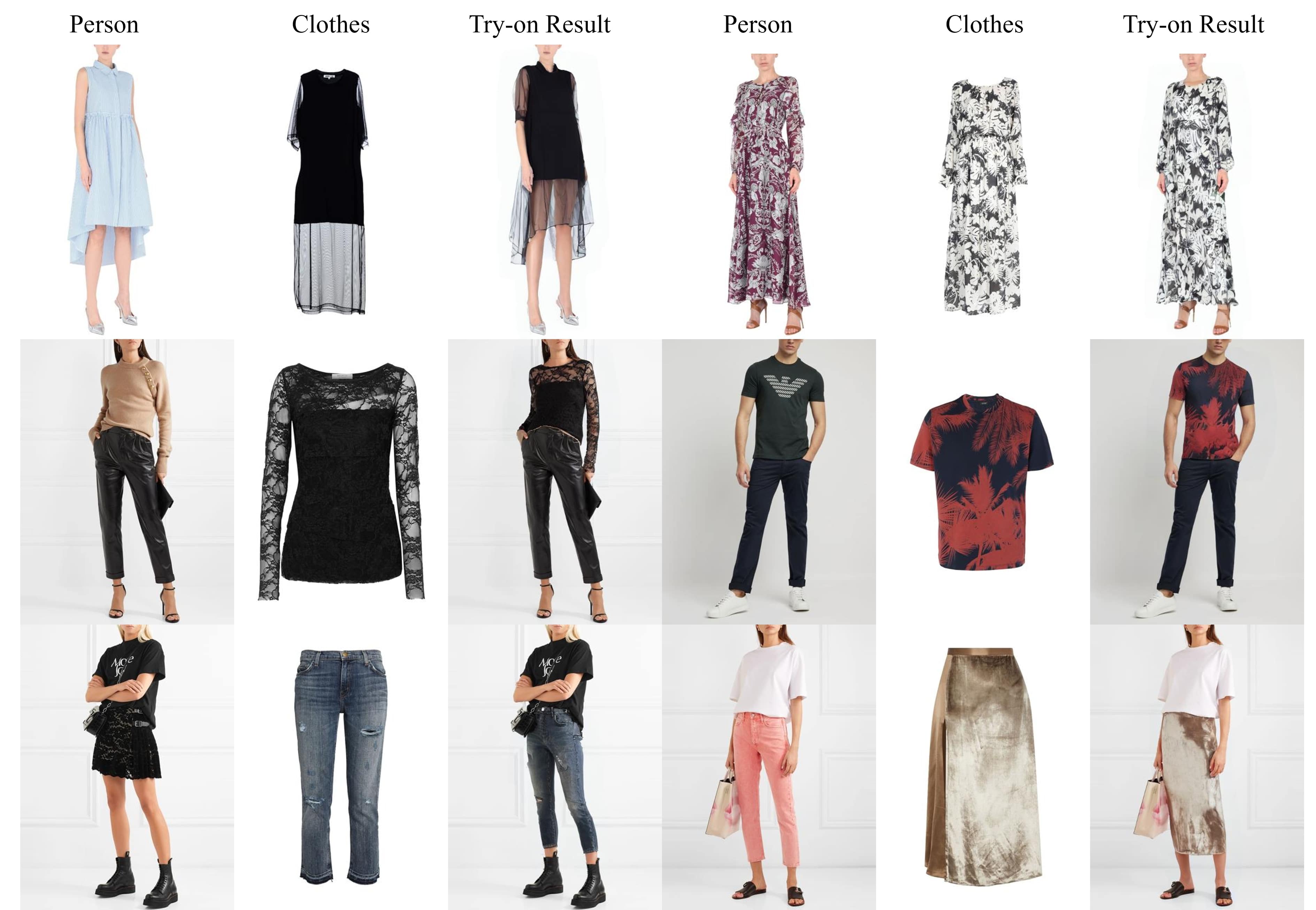}
    \caption{Visualization results on DressCode dataset.}
    \label{fig:dresscode}
\end{figure*}

Similar to VITON-HD~\cite{choi2021viton}, DressCode~\cite{morelli2022dress} is a dataset containing high-quality try-on data pairs, and is consist of three sub-datasets, namely dresses, upper-body and lower-body. Overall, the dataset is composed of 53,795 image pairs: 15,366 pairs for upper-body clothes, 8,951 pairs for lower-body clothes, and 29,478 pairs for dresses. For training, we followed the method of extracting the agnostic mask in \citep{morelli2022dress}, and the rest of the settings were consistent with the setting in VITON-HD. All experiments on the DressCode dataset are performed at $512 \times 384$ resolution.

Table \ref{tab:dresscode} shows the quantitative comparison among PF-AFN~\cite{ge2021parser}, HR-VTON~\cite{lee2022high} and Ours. We measure the LPIPS and FID metrics of the three for paired an unpaired setting respectively. It can be seen that our method achieves the best performance on all three sub-datasets.  Besides that, we visualize the results of our method, as shown in Figure \ref{fig:dresscode}. In the three sub-datasets, our method can achieve realistic and natural try-on results.

\subsection{Results on DeepFashion}
\begin{figure*}
    \centering
    \includegraphics[width=\linewidth]{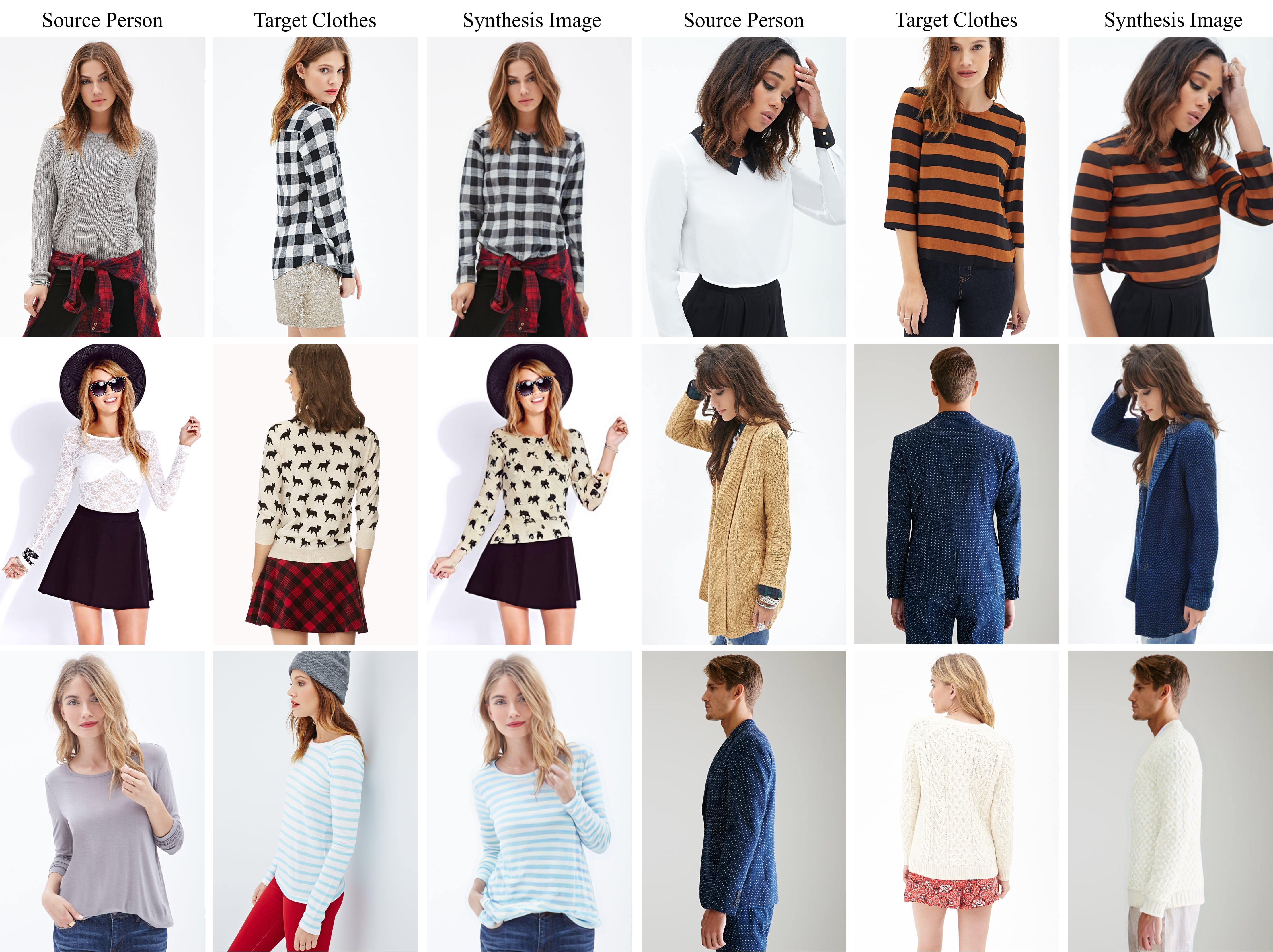}
    \caption{Visualization results on DeepFashion dataset.}
    \label{fig:fashion}
\end{figure*}

\begin{figure*}
    \centering
    \includegraphics[width=0.7\linewidth]{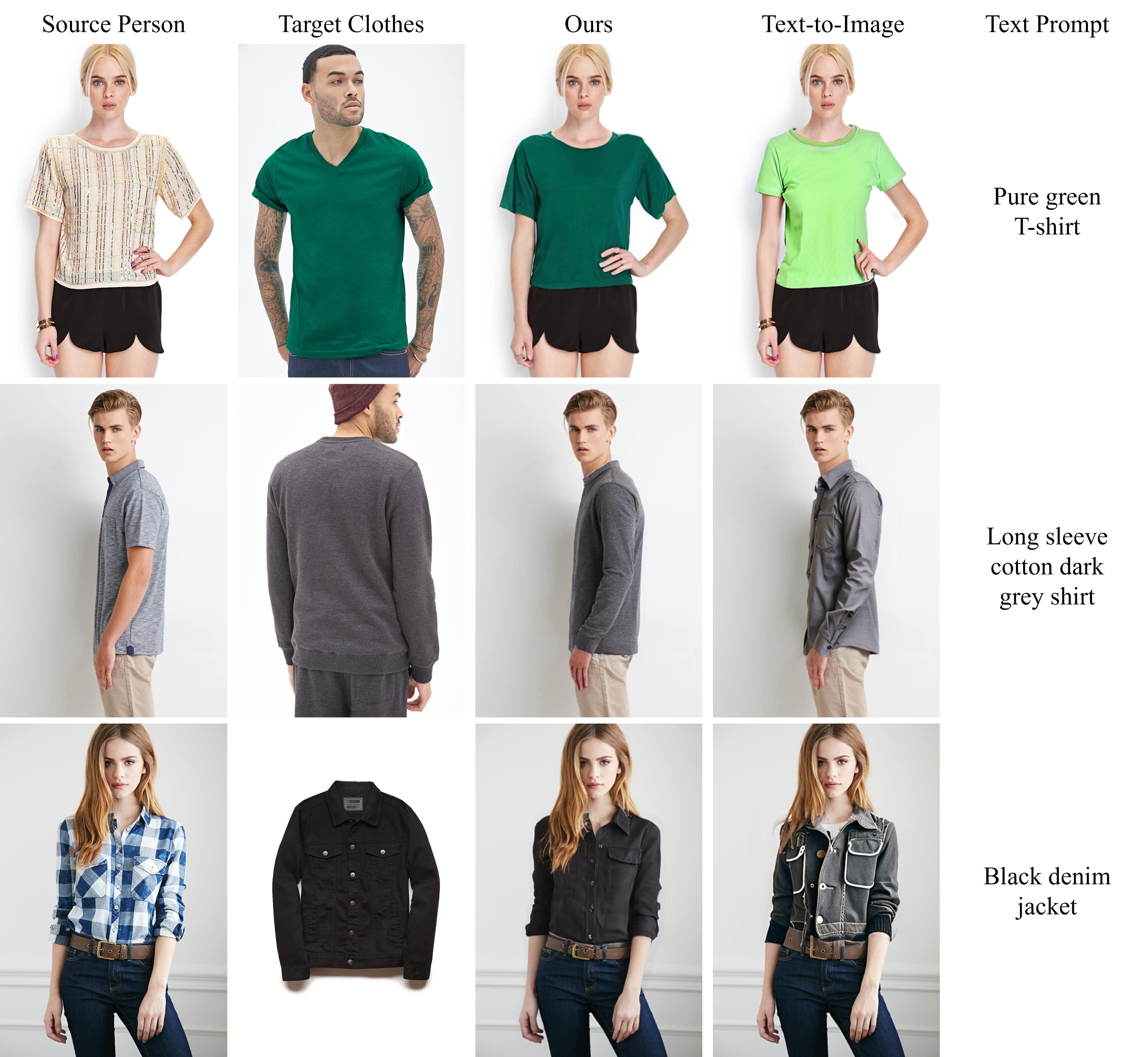}
    \caption{Visual comparison of our method and text-to-image method on DeepFashion dataset.}
    \label{fig:text}
\end{figure*}
In the DeepFashion~\cite{liu2016deepfashion} dataset, our task goal is to transfer the clothes worn by the person in one image to the person in another image. Compared with the previous task background of providing template clothes, this task is undoubtedly more challenging. 

We train our model on the DeepFashion dataset at 512 resolution, and the training process is the same as on the VITON-HD dataset. During the training process, we use two images of the same person in the same dress in different poses as training pairs, and then extract the clothes from one of the image and put it on the person in another image. Following the training/test split used in PATN~\cite{zhu2019progressive} for pose transfer, we first obtained 101,966 data pairs for training. On this basis, we eliminated the data pairs in which the clothes accounted for too little in the image, and finally obtained 51,644 pairs for training.

In Figure \ref{fig:fashion}, we show some visualization results on DeepFashion dataset. It can be seen that even if the clothes are transferred between the person in different poses, our method can  preserve the characteristics of the clothes effectively and generate a realistic composite image.

\subsection{Comparisons to Text-to-Image Approach} \label{sec:text}
Additionally, we experiment with the existing text-to-image inpainting method on DeepFashion to compare with our method. Specifically, we use the pretrained stable diffusion inpainting model~\cite{rombach2022high}, and then use the text description corresponding to the clothes as the condition to generate the final result. Similarly, in the input we will mask the upper half of the human body. The comparison results are shown in Figure \ref{fig:text}. It is clear that utilizing text as the condition alone makes it impossible to recover the qualities of the clothing we need, as the clothing's color, material, and pattern details will vary.

\section{Discussions on Limitations}
\label{sec:limit}

\begin{figure*}
    \centering
    \includegraphics[width=\linewidth]{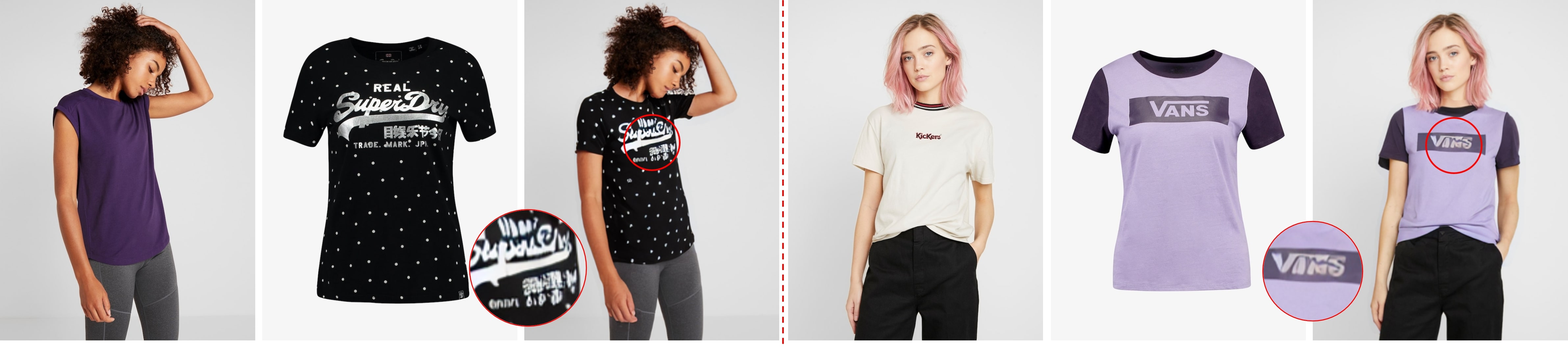}
    \caption{Visualization of our failure cases on VITON-HD dataset. Each sample tuple is the target person, target clothes and composite image from left to right. For these examples, we zoom in for better observation.}
    \label{fig:limit}
\end{figure*}

Despite producing excellent results, our method does not entirely cover all cases. 
As shown in Figure \ref{fig:limit}, we display some less satisfactory composite results. In these two examples, our approach fails to accurately reproduce the clothing patterns. This demonstrates that for some relatively tiny and complex patterns, our method cannot accurately preserve every detail. It is challenging for our method to exactly replicate some little writing on clothing, but for some less strict patterns, the produced results can be fairly consistent. One reason for this could be that the inpainting process takes place in the latent space, which will result in a certain loss, especially for such a small and precise target.
\bibliographystyle{ACM-Reference-Format}
\balance
\bibliography{sample-base}